\documentclass[letterpaper, 10 pt, journal, twoside]{IEEEtran}
\IEEEoverridecommandlockouts 

\usepackage{changes}
\usepackage{amsmath,amsfonts}
\usepackage{algpseudocode}
\usepackage{algorithm}
\usepackage{array}
\usepackage[caption=false,font=normalsize,labelfont=sf,textfont=sf]{subfig}
\usepackage{textcomp}
\usepackage{stfloats}
\usepackage{url}
\usepackage{verbatim}
\usepackage{graphicx}
\usepackage{cite}
\usepackage{hyperref}

\usepackage{siunitx}
\usepackage{caption}
\usepackage{mathtools}
\newcommand*{\tran}{^{\mkern-1.5mu\mathsf{T}}}

\definecolor{changedcol_v2}{rgb}{0,0,0}
\definecolor{changedcol}{rgb}{0,0,0}
\DeclareCaptionFont{ccchngd}{\color{changedcol}}
\DeclareCaptionFont{ccblack}{\color{black}}

\begin{document}

\title{Model Predictive Parkour Control of a Monoped Hopper in Dynamically Changing Environments}

\author{Maximilian Albracht$^{1,2}$, Shivesh Kumar$^{1,3}$, Shubham Vyas$^{1,4}$, Frank Kirchner$^{1,4}$ 

\thanks{Manuscript received: April, 16, 2024; Revised July, 6, 2024; Accepted July, 29, 2024.}
\thanks{This paper was recommended for publication by
Editor Abderrahmane Kheddar upon evaluation of the Associate Editor and
Reviewers’ comments. [Note that the Editor is the Senior Editor who
communicated the decision; this is not necessarily the same as the
Editor-in-Chief.] This work has been supported by the M-RoCK~(FKZ 01IW21002) and VeryHuman~(FKZ 01IW20004) projects funded by the German Aerospace Center (DLR) with federal funds from the Federal Ministry of Education and Research (BMBF) and with project funds from the federal state of Bremen for setting up the Underactuated Robotics Lab (201-342-04-1/2023-4-1).} 
\thanks{$^{1}$Robotics Innovation Center, DFKI GmbH, 28359 Bremen, Germany. \textit{firstname.lastname@dfki.de}}
\thanks{\textcolor{changedcol_v2}{$^{2}$Space Flight Technology, German Space Operations Center, German Aerospace Center (DLR), We{\ss}ling, Germany. \textit{maximilian.albracht@dlr.de}}}
\thanks{$^{3}$Dynamics Division, Department of Mechanics \& Maritime Sciences, Chalmers University of Technology, Gothenburg, Sweden.}
\thanks{$^{4}$AG Robotik, University of Bremen, 28359 Bremen, Germany.}
\thanks{Digital Object Identifier (DOI): see top of this page.}
}



\maketitle

\begin{abstract}
A great advantage of legged robots is their ability to operate on particularly difficult and obstructed terrain, which demands dynamic, robust, and precise movements.
The study of obstacle courses provides invaluable insights into the challenges legged robots face, offering a controlled environment to assess and enhance their capabilities.
Traversing it with a one-legged hopper introduces intricate challenges, such as planning over contacts and dealing with flight phases, which necessitates a sophisticated controller.
A novel model predictive parkour controller\footnote{https://github.com/dfki-ric-underactuated-lab/mppc} is introduced, that finds an optimal path through a real-time changing obstacle course with mixed integer motion planning\footnote{https://www.youtube.com/watch?v=vxFpLKi-pIQ}.
The execution of this optimized path is then achieved through a state machine employing a PD control scheme with feedforward torques, ensuring robust and accurate performance. 
\end{abstract}
\vspace{-0.4cm}
\vspace{-0.4cm}
\section{Introduction}
\IEEEPARstart{L}{egged} robots exhibit significant potential for diverse and challenging environments due to their agility and versatility.
They can navigate through highly variable terrains, leap across large gaps, and traverse obstacles that conventional wheeled robots would fail to overcome.
Potential fields of use for legged robotic systems are among others search and rescue operations, environmental monitoring, and exploration of areas inaccessible to humans. 
In contrast to traditional locomotion methods, where obstacles typically impede progress, \textcolor{changedcol}{human} practitioners of \textcolor{changedcol}{the discipline} parkour (traceurs) perceive obstacles as chances to harness reaction forces through meticulous body control.
There are sophisticated climbing~\cite{Degani2014} and brachiation robots~\cite{Javadi2023, 2024_ricmonk}, hopping robots \cite{Haldane2017} \cite{Ding2017}, quadrupedal robots \cite{Hutter2016} \cite{Bledt2018}, and bipedal humanoid robots \cite{Hubicki2016} \cite{Gong2019} that are capable of navigating harsh environments with high agility.

A particularly difficult pattern of movement for monoped legged robots, the hopping gait, is used to overcome gorges or large hurdles.
\textcolor{changedcol}{It combines highly nonlinear flight phase dynamics along with binary decisions on foot placement during contact with the environment.
Because of this, motion planning for hopping is inherently a mixed integer planning problem. Mixed-integer programming (MIP) has found many use cases in robotics~\cite{Deits2014, Griffin2019, Marcucci2019}.
To study hopping, one-legged robots are convenient, as this is the only gait they are capable of.
A summary of early research on the hopping gait with single-legged robots is provided in \cite{Sayyad2007}.
A widely used model for hopping is the spring-mass model \cite{Blickhan1989}, also known as the Spring Loaded Inverted Pendulum (SLIP).
It is used in various related model-based control approaches to traverse parkour-like terrain \cite{Holmes2006, Arslan2012, Rutschmann2012, Degani2014, Zamani2020}.}
\begin{figure}[t]
	\centering
	\includegraphics[width=0.9\linewidth]{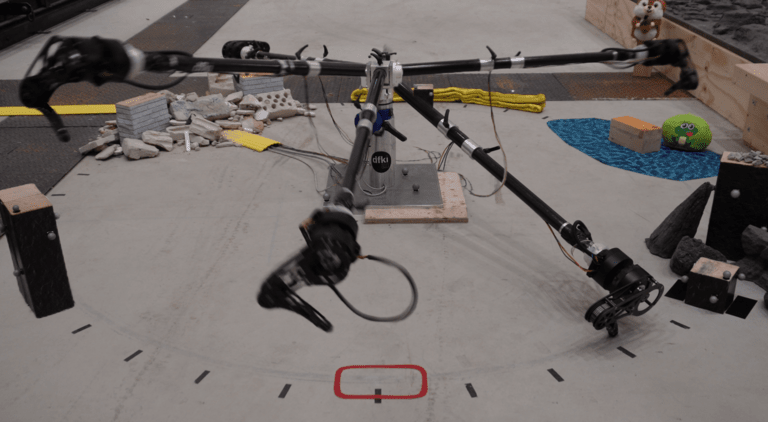}
	\caption{Combined snapshots of a parkour traversal.}
	\label{fig:strike}
	\vspace{-0.6cm}
\end{figure}
\textcolor{changedcol}{\cite{Holmes2006} is a review of the field, while \cite{Arslan2012} provides a deadbeat controller and its region of attraction analysis for wind and sensor noise in simulation with no experimental results. \cite{Rutschmann2012} gives an MPC for hopping with pre-determined foot positions which is also validated in simulation. An inclined slope wall climbing robot with its basic hopping controller is described in \cite{Degani2014}. \cite{Zamani2020} adds the model of the terrain as cubic spline in the cost function of the MPC formulation in simulation thus evading the mixed integer problem. However, this formulation can only avoid danger areas but cannot exploit the terrain in contrast to our work.
The works \cite{Ding2020} and \cite{Ding2021hybrid} formulate the problem either as a mixed integer convex optimization problem or Kinodynamic RRT + trajectory optimization respectively and showcase them on robotic setup with 2-3 jumps. Both works require the solve time in the order of seconds making the planning not suitable for online application. 
\cite{Fey2024} uses MIP as a low-level planner to generate \textcolor{changedcol_v2}{a pre-planned} approximate reference trajectory for hopping with an impulsive stance phase.} 
\textcolor{changedcol}{Learning-based control methods hold promise for parkour with quadruped robots as demonstrated in \cite{Rudin2022} with Proximal Policy Optimization as the Deep Reinforcement Learning (RL) algorithm, in \cite{Cheng2023} with end-to-end RL, in \cite{Zhuang2023} with two stage reinforcement learning, and in \cite{Caluwaerts2023} with a transformer network for locomotion.
However, learning-based approaches typically require a large amount of training data, which can be very resource-intensive.}
In contrast, model-based approaches incorporate a precise understanding of the environment into the control strategy. This allows for robust and adaptable control, even in complex and dynamic environments, without requiring extensive training data.
\textcolor{changedcol}{
As seen in literature, there is a clear lack of a combined planning and control method for a monoped hopping leg which can solve the inherently mixed integer problem \emph{online} thereby taking advantage of the complex terrain (e.g. using one obstacle to jump over another), and adapt to dynamically changing terrain.
}

\textcolor{changedcol}{In this letter, we propose a novel Model Predictive Parkour Control (MPPC) algorithm for impulse planning in parkour hopping that is highly suitable for real-time applications. This allows our hopper to navigate a complex parkour terrain with several obstacles and keep-out zones as compared to the parkour capabilities demonstrated in the literature~\cite{Ding2017, Ding2020, Zamani2020}. 
Furthermore, the real-time computation enables online re-planning that makes the robot adaptable to dynamic complex environments and robust to large disturbances.
The MPPC methodology presented here leverages mixed integer programming and adopts an MPC-like approach that takes impulsive behavior of hopping into account, which involves non-smooth dynamics due to contacts.
Similar to the horizon in an MPC, the MPPC has a lookahead distance which changes as the robot moves through the complex terrain. 
This allows us to model the complexity of the optimization problem as naturally increasing with only number of jumps.
Tests are conducted for different parkour setups and the robot shows very dynamic, robust and adaptable behavior. The source code for the work is open-sourced for easier reproducibility.}
Snapshots of a Parkour completion are displayed in \autoref{fig:strike}. 

\paragraph*{Organization} \autoref{sec:setup} introduces the system setup where we outline the robotic testbench and justify model \textcolor{changedcol}{assumptions.
\autoref{sec:behaviour_gen} describes the core contribution of this paper, the impulse planning with mixed integer optimization in a model predictive control sense.
A behavior state machine to realize the parkour is presented in \autoref{sec:stabilization} using the planned impulses.}
We present the experimental evaluation of multiple parkour traversals and assess disturbance rejection in \autoref{sec:result}.
\autoref{sec:conclusion} concludes the paper and identifies future research directions.

\vspace{-0.3cm}
\section{Sytem Setup}
\label{sec:setup}
\textcolor{changedcol}{The robotic system comprises of a 2 degrees of freedom (DoF) leg connected to a 2 DoF boomstick.
The boomstick contains two rotational joints enabling the leg to undergo passive pitch and yaw rotations. The hip and knee joint of the leg are the active degrees of freedom, equipped with a qdd100 motor from mjbots \cite{qdd100}.} These are quasi-direct drives that allow for high torque-to-mass ratio and back drivability.
Both motors are allocated at the hip, which results in a very low inertia of the leg limbs.
The leg without the support structure shares the exact specifications with the hopping leg discussed in~\cite{Soni2023} and is part of \textit{RealAIGym}~\cite{2022_rss_realaigym}.
Similar robotic test benches have been built and published in \cite{Ding2017}, \cite{Ramos2021}. 
\vspace{-0.1cm}
\subsection{\textcolor{changedcol}{Assumptions}}
\begin{figure}[h!]
\centering
\includegraphics[width=0.98\linewidth]{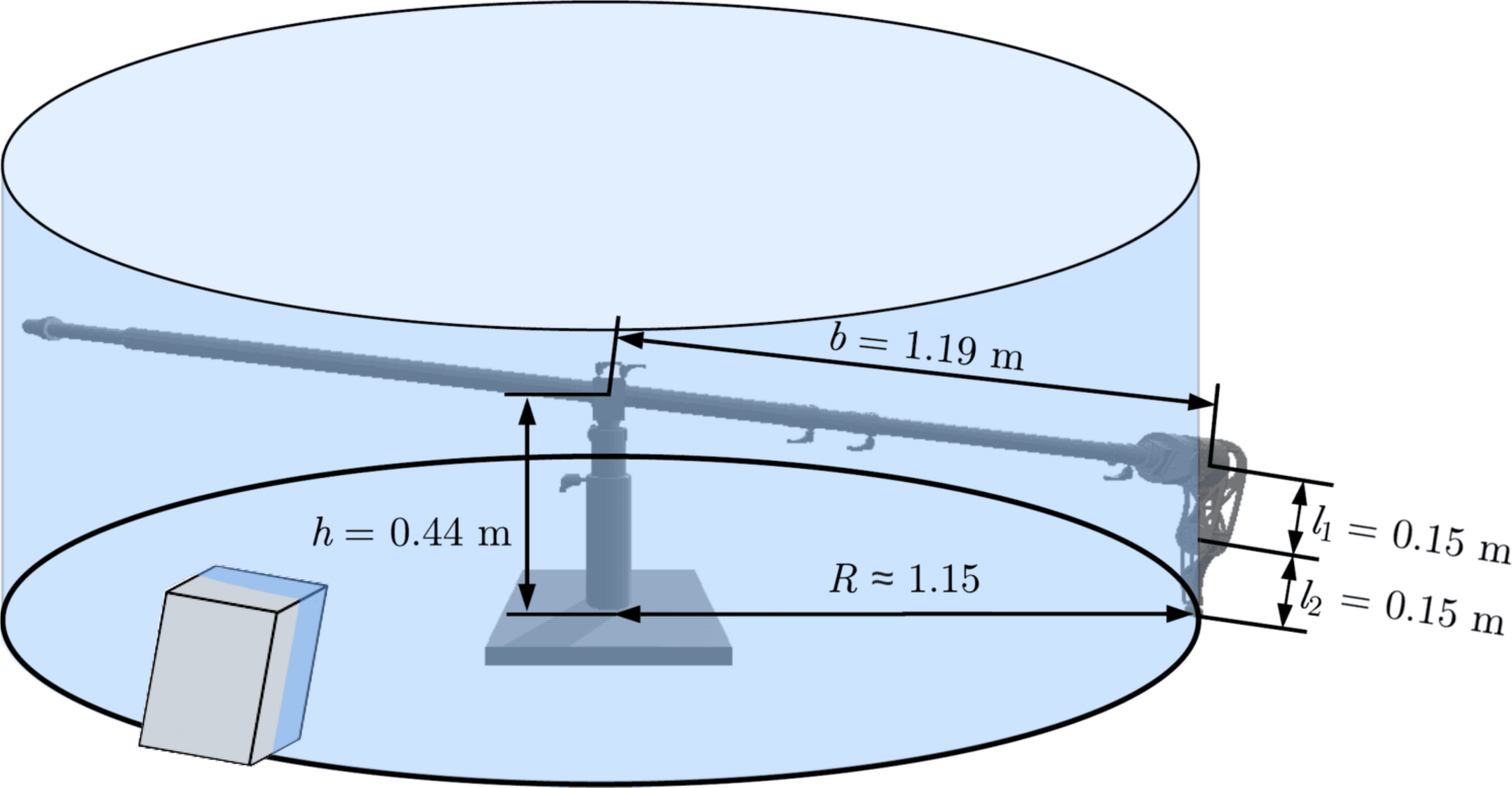}
\caption{Workspace idealization towards a cylindrical surface.}
\label{workspace}
\end{figure}

Because of the much higher mass of the motors compared to the lightweight limbs, it is reasonable to assume the COM at the hip and treat the limbs to be weightless.
Without limits for the four degrees of freedom and the ground boundary, the workspace of the system is a thin-walled hollow sphere.
With the joint limits and a very long rod, the workspace can be approximated with a cylindrical surface as seen in \autoref{workspace}.
For the given dimensions and limits of the system, this representation of the workspace is not exact but is still sufficiently accurate.
Making this assumption makes the control problem far more convenient, reduces computational demands and provides a clearer and easier representation and visualization.
Further, it is assumed that the legs mass distribution is of a point mass located at the hip.
This is appropriate, as $\geq 90 \%$ of the legs mass is lumped at the hip.
Ballistic calculations of the leg are therefore carried out on the two-dimensional plane of the cylindrical surface for a point mass without drag.

\textcolor{changedcol}{For impulse planning, the problem is modeled as time-of-flight optimization. The optimization of the stance phase times are neglected. While in reality they could be important to make the traversal through the parkour course slightly faster, it could pose additional computational demand on the controller.}
\subsection{Parkour Environment}
The surface on which the Hopping Leg is situated is flat and homogeneous. Obstacles come in various dimensions, allowing for a diverse and flexible parkour design.
For simplicity in modeling, the obstacles are limited to a block-like shape resembling a cuboid.
One side of the block must maintain full contact with the ground, resulting in only horizontal and vertical surfaces within the parkour.
The obstacles serve multiple purposes; they can be utilized as platforms for landing or merely as hurdles to be jumped over.
This versatility adds complexity and challenge to the parkour, enabling the Hopping Leg to demonstrate its dynamic and adaptive capabilities during traversal.
Further parkour elements are restricted areas.
They can represent bodies of water, chasms, or arbitrarily designated areas where contact is undesirable in the real world.
Some obstacles are equipped with markers making them traceable objects for the \textit{Vicon} motion capture system \textcolor{changedcol}{which gives robot vision.}
Those obstacles can be added, removed or relocated during the execution, forcing the controller to adapt to environmental changes in real-time.
The objects are tracked in every time step and the parkour model is updated directly before every behavior generation.

\section{\textcolor{changedcol}{Impulse Planning with Mixed Integer Optimization}}
\label{sec:behaviour_gen}
\begin{figure}[t]
\centering
\includegraphics[width=0.85\linewidth]{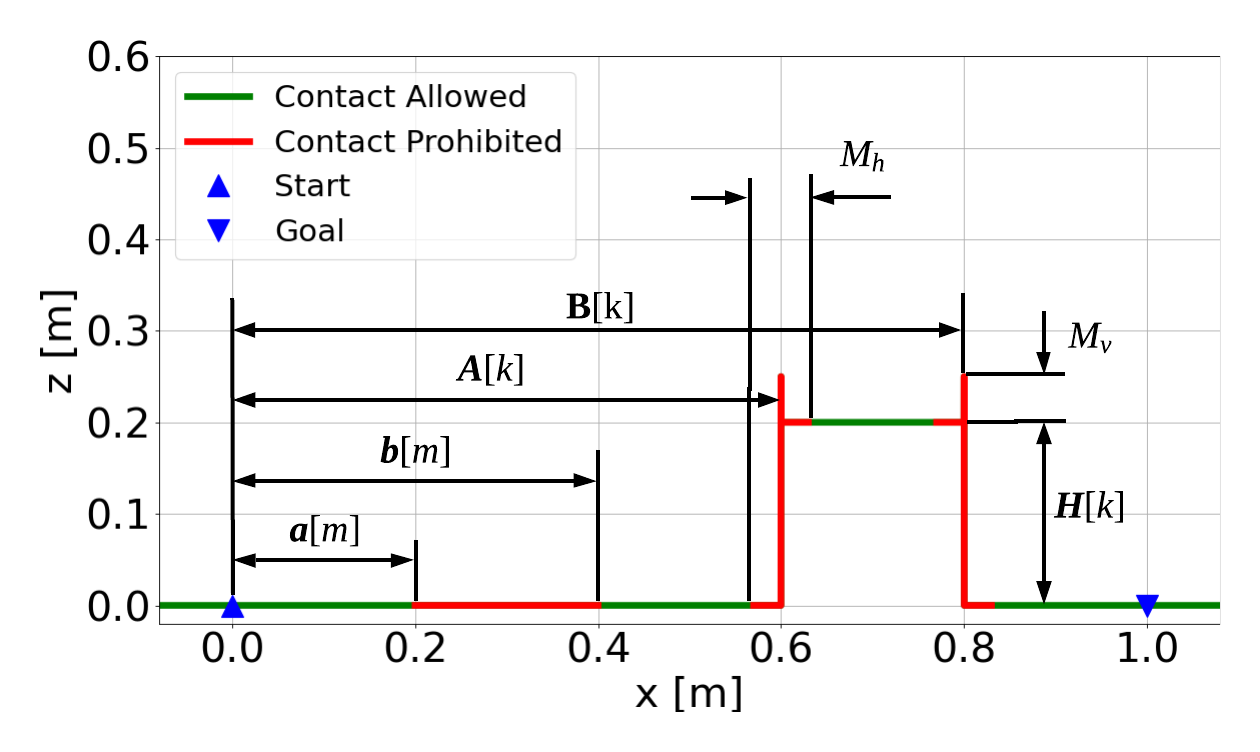}
\caption{Exemplary parkour environment model}
\label{pic:model}
\vspace{-0.3cm}
\end{figure}

\textcolor{changedcol}{For impulse planning, the dynamics of the hopping leg is reduced to a point mass model.}
The point mass is intended to jump to a specific point, with the flexibility to individually adjust parameters for each jump.
Additionally, it must reach this target using a total of $N$ jumps while overcoming $K$ obstacles and navigating through $M$ restricted areas.
\autoref{pic:model} displays a sketch that depicts the mathematical model of a parkour instance with one hurdle and one restricted area graphically. 
The obstacles are modeled into the two-dimensional workspace \textcolor{changedcol}{$\forall k \in \left\{ 1, \ldots K \right\}$} with a start position $\boldsymbol{A}[k]$ and an end position $\boldsymbol{B}[k]$ with a height $\boldsymbol{H}[k]$. 
Additionally, each obstacle inherits a horizontal margin $M_h$ and a vertical margin $M_v$ which are not allowed to come in contact with the leg.
All restricted areas are modeled \textcolor{changedcol}{$\forall m \in \left\{ 1, \ldots M \right\}$} with a start position $\boldsymbol{a}[m]$ and an end position $\boldsymbol{b}[m]$. 
Start and goal coordinates are marked, and possible landing regions are colored in green.
Highlighted in red are regions that are not allowed to get in contact with any part of the leg.

\begin{figure}[t]
\centering
\includegraphics[width=0.85\linewidth]{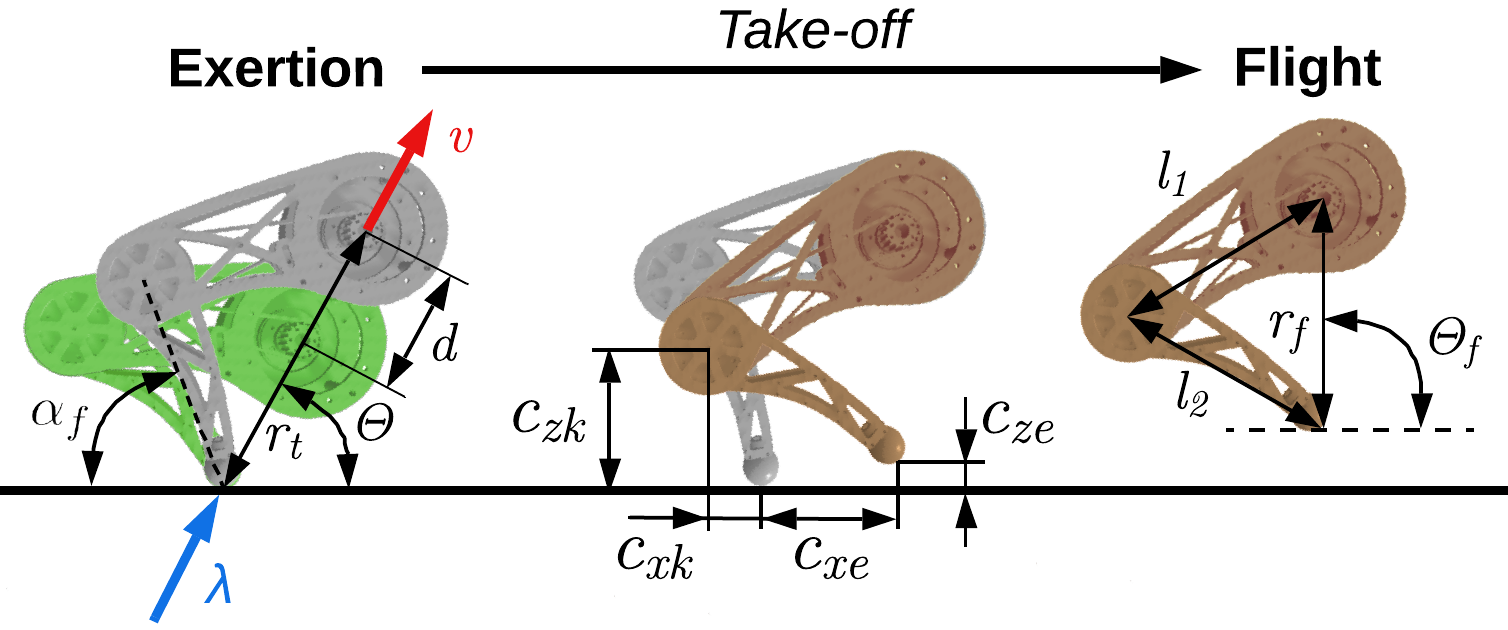}
\caption{\textcolor{changedcol}{Leg orientation shift after take-off.}}
\label{pic:legModel}
\vspace{-0.35cm}
\end{figure}

\textcolor{changedcol}{
\subsection{System Modeling}
The leg moves through different configurations during a jump and the necessary model variables are introduced here.}
After takeoff, the leg moves nearly instantaneously into the flight orientation to prepare for a smooth landing.
The difference between these two orientations acts as \textcolor{changedcol}{a landing} offset to the ballistic trajectory.
\textcolor{changedcol}{Figure \ref{pic:legModel} depicts the leg in the take-off and flight configurations with the corresponding dimensions.}
The foot shifts in the $x$ direction with $\boldsymbol{c_{xe}}[n]$ and in the $z$ direction with $\boldsymbol{c_{ze}}[n]$, depending on the take-off extension $r_{t}$ (distance from foot to hip), the flight extension $r_f$, and the flight angle $\theta_f$ (angle between the ground and the connecting line from foot to hip).
As the knee is a potential source of unintentional contact with obstacles, this offset must also be determined in the $x$ direction $\boldsymbol{c_{xk}}[n]$ and $z$ direction $\boldsymbol{c_{zk}}[n]$.
The knee distances are calculated taking into account the link length of the lower limb $l_2$ and the angle between the lower limb and the ground $\alpha_f$ during flight.
\begin{align}
\boldsymbol{c_{xe}}[n] &= r_{t} \cos{\boldsymbol{\theta}[n]} - r_f \cos{\theta_f} \\
\boldsymbol{c_{ze}}[n] &= r_{t} \sin{\boldsymbol{\theta}[n]} - r_f \sin{\theta_f} \\
\boldsymbol{c_{xk}}[n] &= \boldsymbol{c_{xe}}[n] - l_2 \cos{\alpha_f} \\
\boldsymbol{c_{zk}}[n] &= \boldsymbol{c_{ze}}[n] + l_2 \sin{\alpha_f}
\end{align}

\subsection{Optimization Problem Formulation}
\textcolor{changedcol}{
Parkour impulse planning is formulated as a mixed-integer optimization problem with the goal of minimizing the total duration of flight phases of all jumps. The optimization problem is mathematically formulated as follows:}
\begin{subequations}
\begin{align}
& & & \underset{\boldsymbol{t}, \boldsymbol{v}, \boldsymbol{\theta}, \boldsymbol{\Delta_a}, \boldsymbol{\Delta_b}}{\text{minimize}} \quad \sum_{n=1}^N \boldsymbol{t}[n] \\
& & & \text{subject to} \nonumber \\
& \text{\textcolor{changedcol}{Air-time:}} & & \boldsymbol{t}_\text{min} \leq \boldsymbol{t} \leq \boldsymbol{t}_\text{max} \\
& \text{\textcolor{changedcol}{Take-of velocity:}} & & \boldsymbol{v}_\text{min} \leq \boldsymbol{v} \leq \boldsymbol{v}_\text{max} \\
& \text{\textcolor{changedcol}{Take-of angle:}} & & \boldsymbol{\theta}_\text{min} \leq \boldsymbol{\theta} \leq \boldsymbol{\theta}_\text{max} \\
& \text{\textcolor{changedcol}{Respect dynamics:}} & & d(\boldsymbol{t}, \boldsymbol{v}, \boldsymbol{\theta}) = 0 \label{equ:con:d} \\
& \text{\textcolor{changedcol}{Localize landing point:}} & & \boldsymbol{o_1}(\boldsymbol{t}, \boldsymbol{v}, \boldsymbol{\theta}, \boldsymbol{\Delta_a}, \boldsymbol{\Delta_b}) \leq 0 \label{equ:con:o1} \\
& \text{\textcolor{changedcol}{Respect obstacle height:}} & & \boldsymbol{o_2}(\boldsymbol{t}, \boldsymbol{v}, \boldsymbol{\theta}, \boldsymbol{\Delta_a}, \boldsymbol{\Delta_b}) = 0 \label{equ:con:o2} \\
& \text{\textcolor{changedcol}{Avoid landing on edges:}} & & \boldsymbol{o_3}(\boldsymbol{t}, \boldsymbol{v}, \boldsymbol{\theta}) \leq 0 \label{equ:con:o3} \\
& \text{\textcolor{changedcol}{Avoid collision:}} & & \boldsymbol{a}(\boldsymbol{t}, \boldsymbol{v}, \boldsymbol{\theta}, \boldsymbol{\Delta_a}, \boldsymbol{\Delta_b}) \leq 0 \label{equ:con:c}\\
& \text{\textcolor{changedcol}{Avoid restricted areas:}} & & \boldsymbol{r}(\boldsymbol{t}, \boldsymbol{v}, \boldsymbol{\theta}) \leq 0 \label{equ:con:r}
\end{align}
\label{eqn_miopt}
\end{subequations}


\textcolor{changedcol}{
The ballistic trajectory for $n^\text{th}$ jump with $n \in \left\{ 1, \ldots N \right\}$ is fully defined by the following three decision variables: $\boldsymbol{t}[n] \equiv$ air-time of jump, $\boldsymbol{v}[n] \equiv$ take-off velocity of jump and $\boldsymbol{\theta}[n] \equiv$ take-off angle of jump.
}
\textcolor{changedcol}{Further, for every $n^\text{th}$ jump and $k^\text{th}$ obstacle, we define two binary decision variables $\boldsymbol{\Delta_a}[n, k] \in [0,1] \equiv$ which decides whether the foot is in front of the obstacle $k$ passed in jump $n$ and $\boldsymbol{\Delta_b}[n, k] \in [0,1] \equiv$ which decides whether the foot is in back of the obstacle $k$ passed in jump $n$.
}
All constraints above as well as all functions and equations stated in the following are valid \textcolor{changedcol}{$\forall n \in \left\{ 1, \ldots N \right\}$}, \textcolor{changedcol}{$\forall k \in \left\{ 1, \ldots K \right\}$} and \textcolor{changedcol}{$\forall m \in \left\{ 1, \ldots M \right\}$} if dependent on $n$, $k$, $m$ respectively. 
\textcolor{changedcol}{The formulation is devised with Drake \cite{DRAKE} and is solved by the Snopt solver \cite{Gill2005} with the branch-and-bound method for a total of $3N$ continuous and $2NM$ binary decision variables.}

\subsubsection{Dynamics}
The leg's motion is achieved through jumps that follow ballistic trajectories.
To model the dynamics, the velocity components $\boldsymbol{v_x}$ and $\boldsymbol{v_z}$ for the jumps are initially determined.
\begin{align}
\boldsymbol{v_x}[n] &= \boldsymbol{v}[n] \cos{\boldsymbol{\theta}[n]} \\
\boldsymbol{v_z}[n] &= \boldsymbol{v}[n] \sin{\boldsymbol{\theta}[n]}
\end{align}
The contact points $\boldsymbol{x}[n]$ and $\boldsymbol{z}[n]$ are determined from the initial positions $\boldsymbol{x}[0] = x_s$ and \textcolor{changedcol}{$\boldsymbol{z}[0] = z_s$}. 
\begin{align}
\boldsymbol{x}[n] &= \boldsymbol{x}[n-1] + \boldsymbol{c_{xe}}[n] + \boldsymbol{v_x}[n] \boldsymbol{t}[n] \\
\boldsymbol{z}[n] &= \boldsymbol{z}[n-1] + \boldsymbol{c_{ze}}[n] + \boldsymbol{v_z}[n] \boldsymbol{t}[n] - \frac{1}{2} g \left( \boldsymbol{t}[n] \right)^2
\end{align}
It should be noted that the leg's movement is not integrated step by step with a specific time interval, but only one timestep per trajectory is used.
This approach is reasonable as the ballistic trajectory of the COM cannot be changed whilst the leg is not in contact with the ground.
The objective is to land on the target position $x_t$ with the last jump.
This is introduced as constraint \textcolor{changedcol}{$d(\boldsymbol{t}, \boldsymbol{v}, \boldsymbol{\theta}) = 0$} (\ref{equ:con:d}) with function $d(\boldsymbol{t}, \boldsymbol{v}, \boldsymbol{\theta})$ defined as:
\begin{equation}
d(\boldsymbol{t}, \boldsymbol{v}, \boldsymbol{\theta}) = x_t - \boldsymbol{x}[N]
\end{equation} 
This constraint is of particular importance as it gives the overall goal of the parkour traversal.
Note, that there is no goal coordinate in the $z$ direction, as it is indirectly enforced because every $x$ coordinate is mapped to a distinct $z$ coordinate during the parkour implementation.

\subsubsection{Obstacles}
Obstacles are modeled with the parameters $\boldsymbol{A}, \boldsymbol{B}, \boldsymbol{H}$ as displayed in \autoref{pic:model}.
The introduced binary variables serve the purpose of modeling the discontinuous function of the parkour by enabling a mapping between positions in the $x$ and $z$ directions of the parkour.
To achieve this, the binary variable $\boldsymbol{\Delta_a}[n, k]$ is designed to always be $0$ if the $x$ position of the current jump $x[n]$ is less than the front position of the corresponding obstacle $\boldsymbol{A}[k]$, and $1$ if it is greater.
Similarly, for the binary decision variable $\boldsymbol{\Delta_b}[n, k]$, the same logic applies, but for the obstacle's back position $\boldsymbol{B}[k]$.
This is enforced by constraint \textcolor{changedcol}{$\boldsymbol{o_1}(\boldsymbol{t}, \boldsymbol{v}, \boldsymbol{\theta}, \boldsymbol{\Delta_a}, \boldsymbol{\Delta_b}) \leq 0$} (\ref{equ:con:o1}) with function $\boldsymbol{o_1}(\boldsymbol{t}, \boldsymbol{v}, \boldsymbol{\theta}, \boldsymbol{\Delta_a}, \boldsymbol{\Delta_b})$ defined as:
\begin{equation*}
\boldsymbol{o_1}(\boldsymbol{t}, \boldsymbol{v}, \boldsymbol{\theta}, \boldsymbol{\Delta_a}, \boldsymbol{\Delta_b}) = \left( \begin{matrix} \boldsymbol{\Delta_a}[n, k] \left( \boldsymbol{A}[k] - \boldsymbol{x}[n] \right) \\
\left( 1 - \boldsymbol{\Delta_a}[n, k] \right) \left( \boldsymbol{x}[n] - \boldsymbol{A}[k] \right) \\
\boldsymbol{\Delta_b}[n, k] \left( \boldsymbol{B}[k] - \boldsymbol{x}[n] \right) \\
\left( 1 - \boldsymbol{\Delta_b}[n, k] \right) \left( \boldsymbol{x}[n] - \boldsymbol{B}[k] \right) \end{matrix} \right)
\end{equation*}
Constraint (\ref{equ:con:o1}) ensures that the respective variable must be \num{0} when the reference position is not surpassed and becomes \num{1} after surpassing the reference position.
With this framework in place, the subsequent constraint \textcolor{changedcol}{$\boldsymbol{o_2}(\boldsymbol{t}, \boldsymbol{v}, \boldsymbol{\theta}, \boldsymbol{\Delta_a}, \boldsymbol{\Delta_b}) = 0$} (\ref{equ:con:o2}) is designed to align the landing height with the obstacle heights $\boldsymbol{H}[k]$ within the parkour.
Function $\boldsymbol{o_2}(\boldsymbol{t}, \boldsymbol{v}, \boldsymbol{\theta}, \boldsymbol{\Delta_a}, \boldsymbol{\Delta_b})$ is therefore defined as:
\begin{equation*}
\boldsymbol{o_2}(\boldsymbol{t}, \boldsymbol{v}, \boldsymbol{\theta}, \boldsymbol{\Delta_a}, \boldsymbol{\Delta_b}) = \boldsymbol{z}[n] - \sum_{k=1}^{K} \boldsymbol{H}[k] \left( \boldsymbol{\Delta_a}[n, k] - \boldsymbol{\Delta_b}[n, k] \right)
\end{equation*}
When the binary variables for the obstacles hold the same value, this results in a value of $0$ for the default height without obstacles.
This occurs when the landing position in the $x$ direction of the current jump is either entirely before or after an obstacle.
Only in cases where $x[n], \hspace{0.5em} \textcolor{changedcol}{\forall n \in \left\{ 1, \ldots N \right\}}$ lies between the front and back position of an obstacle, must the landing height $z[n], \hspace{0.5em} \textcolor{changedcol}{\forall n \in \left\{ 1, \ldots N \right\}}$. 
In theory, any point on an obstacle could be considered a landing point.
However, practical considerations make corner points unfavorable choices.
Small deviations from the planned position could result in completely missing an obstacle.
Therefore, constraint \textcolor{changedcol}{$\boldsymbol{o_3}(\boldsymbol{t}, \boldsymbol{v}, \boldsymbol{\theta}) \leq 0$} (\ref{equ:con:o3}) is introduced to prevent landing points that are too close to the edges.
The horizontal margin $M_h$ is introduced and function $\boldsymbol{o_3}(\boldsymbol{t}, \boldsymbol{v}, \boldsymbol{\theta})$ is defined as:
\begin{equation}
\boldsymbol{o_3} = \frac{M_h}{2} - \left| \boldsymbol{x}[n] - \left( \begin{matrix} \boldsymbol{A}[k] \\ \boldsymbol{B}[k] \end{matrix} \right) \right|
\end{equation}
With constraint (\ref{equ:con:o3}), landing restrictions are introduced for each obstacle at the front and back points, which extend into and away from an obstacle with half the horizontal margin.

\subsubsection{Collision Avoidance}
Collision avoidance is addressed through a method that examines critical points exclusively.
The collision point for the obstacle front is the foot.
By ensuring that the foot avoids collision with the obstacle front, the rest of the hopping leg is guaranteed not to collide with it.
To establish this, a correlation between the front $x$ position of the obstacle and the $z$ position of the foot during each jump is required.
This is achieved by describing the ballistic trajectories of all jumps as functions and evaluating them at the points $\boldsymbol{A}[k]$.
The corresponding time $t_a[n, k]$ at which the function is at position $\boldsymbol{A}[k]$ is determined.
\begin{align}
t_a[n, k] = \frac{\boldsymbol{A}[k] - \boldsymbol{x}[n-1] - \boldsymbol{c_{xe}}[n]}{\boldsymbol{v_x}[n]}
\end{align}
The $z$ position of the foot on the obstacle fronts $\boldsymbol{z_{ae}}[n, k]$ is calculated for all jumps
\begin{align*}
\boldsymbol{z_{ae}}[n, k] = \boldsymbol{z}[n-1] + \boldsymbol{c_{ze}}[n] + \boldsymbol{v_z}[n] t_a[n, k] - \frac{1}{2} g t_a[n, k]^2.
\end{align*}
During traversal over the obstacle's back, collision assessment involves the knee.
For each jump, the time $t_b$ at which the knee is at an obstacle's back $\boldsymbol{B}[k]$ is determined.
\begin{align}
t_b[n, k] = \frac{\boldsymbol{B}[k] - \boldsymbol{x}[n-1] - \boldsymbol{c_{xk}}[n]}{\boldsymbol{v_x}[n]}
\end{align}
A correlation is established between the obstacle's back position $\boldsymbol{B}[k]$ and the $z$ position of each ballistic trajectory function for the knee $\boldsymbol{z_{bk}}[n, k]$ during each jump
\begin{align*}
\boldsymbol{z_{bk}}[n, k] = \boldsymbol{z}[n-1] + \boldsymbol{c_{zk}}[n] + \boldsymbol{v_z}[n] t_b[n, k] - \frac{1}{2} g t_b[n, k] ^2.
\end{align*}
To ensure that no collision occurs, a vertical margin $M_v$ is introduced for the obstacles.
For each jump, it must be ensured that the $z$ position of the foot/knee, when surpassing the obstacle's front/back, is sufficiently high.
This approach also ensures that the landing angles cannot become excessively shallow and that the leg is always descending upon landing.
It is crucial for the feasibility of the trajectory, as it guarantees sufficient friction in the horizontal direction while landing.
The constraint \textcolor{changedcol}{$\boldsymbol{a}(\boldsymbol{t}, \boldsymbol{v}, \boldsymbol{\theta}, \boldsymbol{\Delta_a}, \boldsymbol{\Delta_b}) \leq 0$} (\ref{equ:con:c}) ensures that the vertical margin is respected.
Function $\boldsymbol{a}(\boldsymbol{t}, \boldsymbol{v}, \boldsymbol{\theta}, \boldsymbol{\Delta_a}, \boldsymbol{\Delta_b})$ is defined with $\boldsymbol{\Delta_a}[0, k] = 0$ and $\boldsymbol{\Delta_b}[0, k] = 0$ as:
\begin{align}
&\boldsymbol{a}(\boldsymbol{t}, \boldsymbol{v}, \boldsymbol{\theta}, \boldsymbol{\Delta_a}, \boldsymbol{\Delta_b}) = \nonumber \\
&\left( \begin{matrix} \left( \boldsymbol{H}[k] + M_v - \boldsymbol{z_{ae}}[n] \right) \left( \boldsymbol{\Delta_a}[n, k] - \boldsymbol{\Delta_a}[n-1, k] \right) \\
\left( \boldsymbol{H}[k] + M_v - \boldsymbol{z_{bk}}[n] \right) \left( \boldsymbol{\Delta_b}[n, k] - \boldsymbol{\Delta_b}[n-1, k] \right) \end{matrix} \right)
\end{align}
By consulting the binary variables, only the trajectories of the jumps that skip the corresponding obstacle front or back are checked for sufficiently high clearance.

\subsubsection{Restricted Areas}
To account for restricted areas, constraint \textcolor{changedcol}{$\boldsymbol{r}(\boldsymbol{t}, \boldsymbol{v}, \boldsymbol{\theta}) \leq 0$} (\ref{equ:con:r}) is imposed to narrow down the acceptable range of values for $\boldsymbol{x}$.
The starting point $\boldsymbol{a}[m]$ as well as the endpoint $b[m]$ of all $M$ restricted areas are taken into account in function $\boldsymbol{r}(\boldsymbol{t}, \boldsymbol{v}, \boldsymbol{\theta})$, which is defined as:
\begin{equation}
\boldsymbol{r}(\boldsymbol{t}, \boldsymbol{v}, \boldsymbol{\theta}) = \frac{\boldsymbol{b}[m] - \boldsymbol{a}[m]}{2} - \left| \boldsymbol{x}[n] - \frac{\boldsymbol{a}[m] + \boldsymbol{b}[m]}{2} \right|
\end{equation}

\subsubsection{Initial Guess}
The optimization problem is then solved with an educated initial guess $\boldsymbol{t_i}, \boldsymbol{v_i}, \boldsymbol{\theta_i} = f(N, v_{max}, \theta_{min})$.
The computed initial guess is the solution for a parkour without obstacles and restricted areas, in which each jump is executed with $\theta_{min}$ and the same take-off velocity.
For the binary decision variables $\boldsymbol{\Delta_a}[n, k]$ and $\boldsymbol{\Delta_b}[n, k]$ the initial value is therefore left at $0$ \textcolor{changedcol}{$\forall n \in \left\{ 1, \ldots N \right\}$}, \textcolor{changedcol}{$\forall k \in \left\{ 1, \ldots K \right\}$}. 

\vspace{-0.1cm}
\subsection{Model Predictive Parkour Control Adaptation}
\label{sec:mpc}
\textcolor{changedcol}{
The optimization problem is integrated into a receding horizon controller as shown in Algorithm \autoref{alg:mpc}.
For each jump starting at position $x_s$, the behavior is generated for a segment of the course covered under a lookahead distance $p$ (which must be selected a priori), followed by the execution of the first jump of the solution.
Thus, the hopping leg is supposed to always end on target position $x_t$, calculated as the minimum of the sum $p+x_s$ and the final goal position $x_g$.
We consider the environment model as $\mathcal{E} = \{\boldsymbol{A}, \boldsymbol{B}, \boldsymbol{H},\boldsymbol{a}, \boldsymbol{b}\}$ consisting of all the obstacles and restricted areas in between the starting and target positions as shown in Fig.~\ref{pic:model}.
The initial height $z_s$ is determined using the function $\text{locate}$, by evaluating the environment model $\mathcal{E}$ at the initial position $x_s$.
The solver then tries to solve the optimization problem with a fixed number of jumps $N \in \mathbb{N}$.
The function $\text{jumps}()$ provides a physics-informed initial guess with a range of minimum to maximum number of jumps $\mathcal{N} = \{N_\text{min}, N_\text{max}\}$ required to complete the parkour task by considering a flat terrain without any obstacles. 
The prediction horizon $N$ is adapted in the for loop based on feasibility.
If no solution could be obtained the solver is given an updated problem. The first jump parameters from the optimal jump sequence are thus the output of MPPC algorithm.
}
\vspace{-0.1cm}
\begin{algorithm}
\caption{\textcolor{changedcol}{MPPC Psuedocode}}\label{alg:mpc}
\begin{algorithmic}[1]
\Function{MPPC}{$\mathcal{E}, x_s, x_g, p, v_\text{min}, v_\text{max}, \theta_\text{min}$}
\State $x_t \gets \min{(p + x_s, x_g)}$
\Comment{Target position}
\State $z_s \gets \text{locate}(\mathcal{E}, x_s)$
\Comment{Starting Height}
\State $\mathcal{N} \gets \text{jumps}(x_s, x_t, p_h, v_{min}, v_{max}, \theta_{min})$
\Comment{Guess Prediction Horizon}
\For{$\textbf{each}\ N\ \in\ \mathcal{N}$}
\State $\boldsymbol{\theta},\boldsymbol{v} \gets miopt(\mathcal{E}, N, x_s, x_t, z_s)$
\Comment{Solve Eqn.~\ref{eqn_miopt}}
\If{$\boldsymbol{\theta},\boldsymbol{v}$ is not \textbf{None}}
\State \textbf{return} $\boldsymbol{\theta},\boldsymbol{v}$
\EndIf
\EndFor
\State $\theta \gets \boldsymbol{\theta}[1], v \gets \boldsymbol{v}[1]$ 
\Comment{First jump in the optimal seq.}
\State \textbf{return} $\theta,v$ 
\EndFunction
\end{algorithmic}
\end{algorithm}

\section{\textcolor{changedcol}{Behavior State Machine for Parkour}}
\label{sec:stabilization}
\begin{figure}[h!]
\centering
\includegraphics[width=0.85\linewidth]{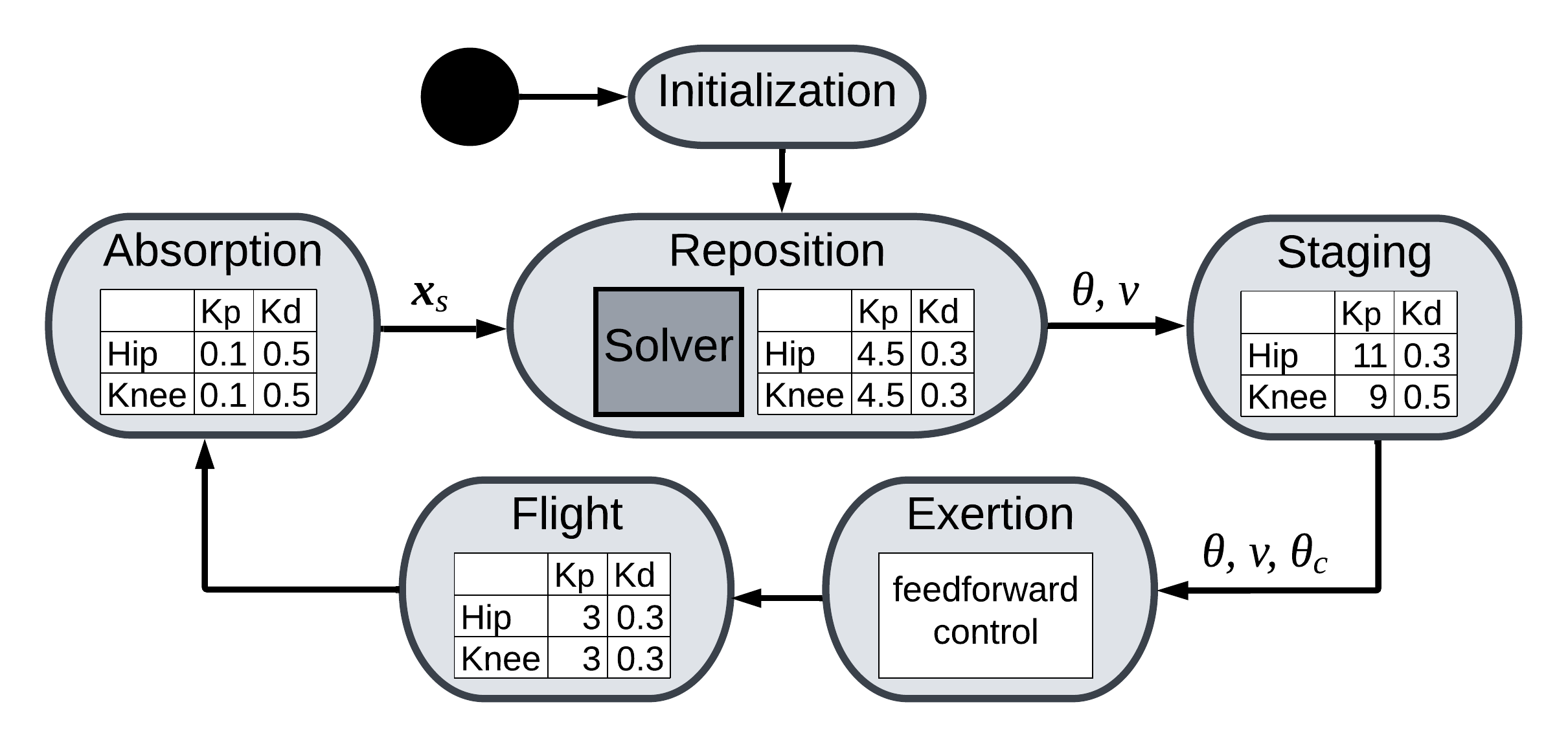}
\caption{\textcolor{changedcol}{State machine to stabilize the behavior}}
\label{statemachine}
\end{figure}

The generated behavior is applied to the real robot with a state machine.
A jump is divided into five phases, each with a specific function to attain smooth jumping behavior.
During each phase, the leg is controlled using PD control.
The \textit{Flight} phase is utilized to align the leg for landing, and in the \textit{Absorption} phase, a significant portion of kinetic energy during landing is dissipated.
In the \textit{Reposition} phase, a new trajectory is generated using the MPPC, and the leg is transitioned into the launch position.
To achieve a precise jump, the foot and hip are made collinear with the force vector.
In an idealized scenario, this would result in the exclusive utilization of the knee motor for force application.
The \textit{Staging} phase is then employed to maintain the leg more accurately in the correct orientation with increased P-gains.
\autoref{statemachine} illustrates the state machine employed to stabilize the behavior \textcolor{changedcol}{with the corresponding \textcolor{changedcol_v2}{joint-space} PD gains}.
The \textit{Initialization} phase is solely used to slowly bring the leg from its initial position into a more crouched position.
Appropriate PD control gains for all states are determined experimentally and ensure a fluid and precise motion.
Jumps are initiated in the \textit{Exertion} phase by pushing off the ground with the foot. 
To initiate the planned jump, torques are calculated and applied using feedforward control during the \textit{Exertion} phase.
\textcolor{changedcol}{Figure \ref{pic:legModel} depicts the leg model in the exertion phase including the dimensions used in the following.}
The behavior generation from \autoref{sec:behaviour_gen} yields the velocity vector $\boldsymbol{v}$ and takeoff angle $\theta$ for the intended jump.
The takeoff angle and leg extension are accepted in a range to facilitate a dynamic transition between phases.
Thus, the actual takeoff velocity $v_c$ is adjusted based on the current takeoff angle $\theta_c$.
\begin{equation}
\boldsymbol{v_{c}} = \sqrt{\boldsymbol{v}^2 \frac{\sin{2 \theta}}{\sin{2 \theta_c}}}
\end{equation}
The exertion distance $d$ is computed from the current leg extension $r_c$ and the planned leg extension at takeoff $r_t$ as $d = r_t - r_c$. Consequently, the constant force $\boldsymbol{\lambda}$ required over the exertion distance is determined in dependence on the leg mass $m$ with 
$
\boldsymbol{\lambda} = \frac{m \textcolor{changedcol}{\boldsymbol{v_c}}^2}{2 d}.
$
This force, also known as the ground reaction force, is utilized to calculate the necessary joint torques using $\boldsymbol{\tau} = \boldsymbol{J}\tran \boldsymbol{\lambda}$.
This calculation occurs at each time step during the \textit{Exertion} phase, as the Jacobian is a function of the changing joint positions.

\section{Experimental Results}
\label{sec:result}
The MPPC controller is used to move the hopping leg through a variety of arbitrary set-up obstacle courses in which the obstacles can appear, disappear or move in real time.
To perceive moving obstacles, their position and orientation are tracked with a motion capture system from \textit{Vicon}.
We provide video footage of the experiments at \footnote{https://www.youtube.com/watch?v=vxFpLKi-pIQ} and open source the code at \footnote{https://github.com/dfki-ric-underactuated-lab/mppc}.
The backflip in the last jump of each experiment is performed as a visual treat only.

\subsection{Static Environment}
\begin{figure}[h!]
\centering
\includegraphics[width=0.85\linewidth]{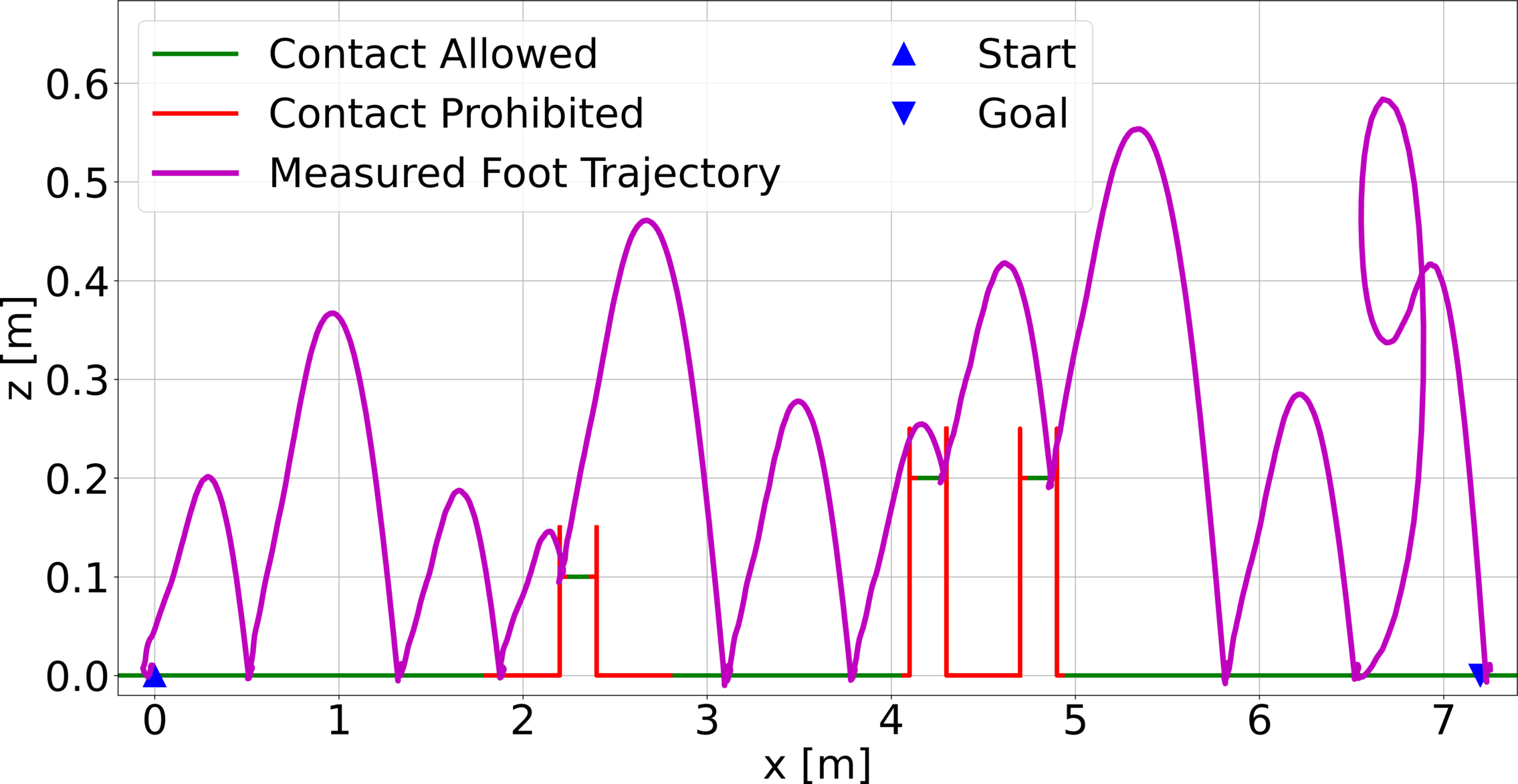}
\caption{Measured foot trajectory for a static obstacle course}
\label{fig:result_1}
\end{figure}

A baseline is established with an environment that features static objects only.
The goal is to perform a full rotation (\SI{7.2}{\metre}) by passing three obstacles and three restricted areas.
The robot achieves this goal within eleven jumps by landing on two obstacles.
A visualization of the environment, along with the measured trajectory of the foot, is depicted in \autoref{fig:result_1}.
Encoders at the unactuated joints at the base help in providing full state feedback allowing the foot position to be computed in real time during experiments.
The measured foot trajectory in the two-dimensional plot is obtained by projection of the actual foot position onto the cylindrical surface of the idealized workspace. \\
\vspace{-0.4cm}
\subsection{Dynamic Environment}
\begin{figure}[h!]
\centering
\includegraphics[width=0.85\linewidth]{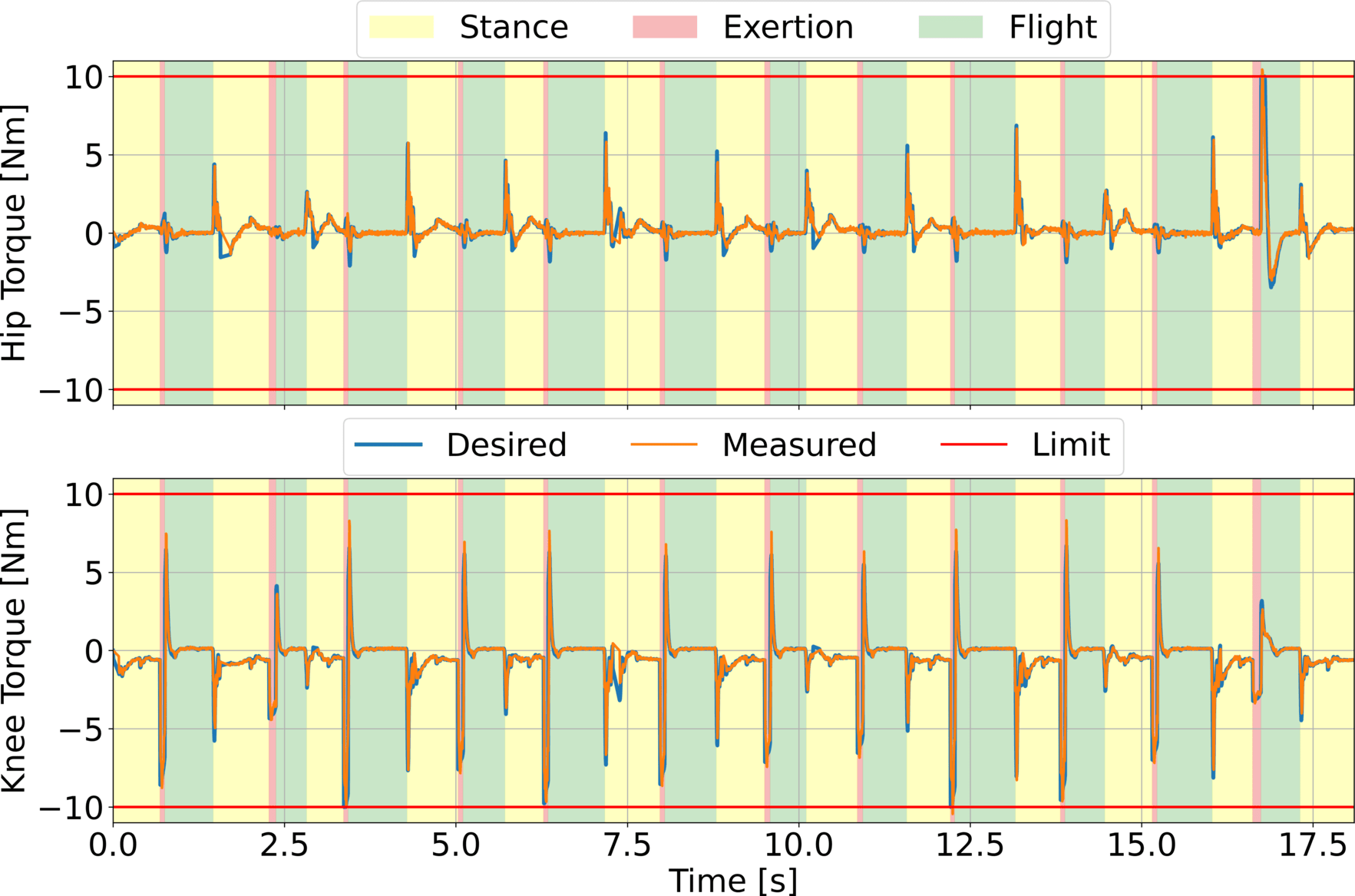}
\caption{Desired and measured joint torque for active DoF}
\label{pic:torque}
\vspace{-0.7cm}
\end{figure}

The dynamic environment consists of three fixed obstacles, four restricted areas and two tracked obstacles that are replaced after the first traversal of the robot.
The tracked obstacles are placed randomly on the circular track during the experiment to show the controller's capability to adapt to a changing environment in real time.
This is realized by sliding the first obstacle across the parkour after being passed to its second position, and by placing the second obstacle in a random position behind the last fixed obstacle during experiment execution.
In total, the robot jumps over seven obstacles and four restricted areas in twelve jumps and uses five obstacles as a platform.
The two dynamic obstacles are tracked in every timestep and are modeled on the two-dimensional workspace for the MPPC before every behavior generation.
Figure \ref{fig:result_2} shows a full traversal along with the section-wise planned trajectories.
{\begingroup
\begin{table*}[htbp]
\setlength{\tabcolsep}{1pt}
\renewcommand{\arraystretch}{0.3}	
\newcommand{\xscale}{0.24\linewidth}
\centering
\begin{tabular}{cccc}
\includegraphics[width=\xscale]{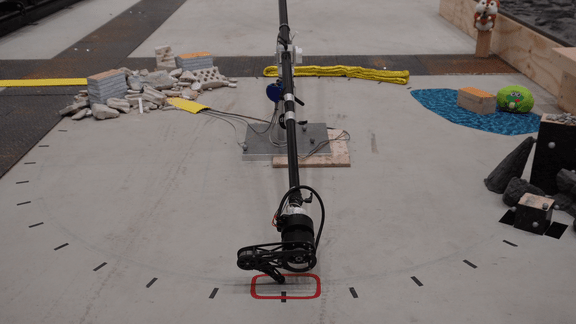}&
\includegraphics[width=\xscale]{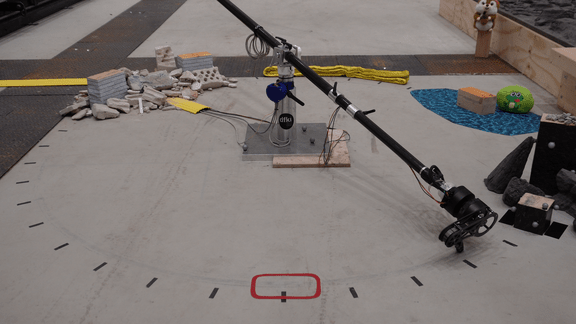}&
\includegraphics[width=\xscale]{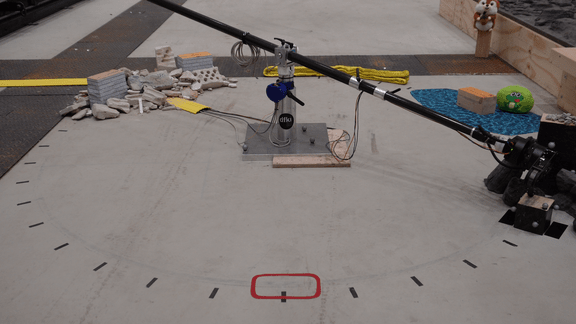}&
\includegraphics[width=\xscale]{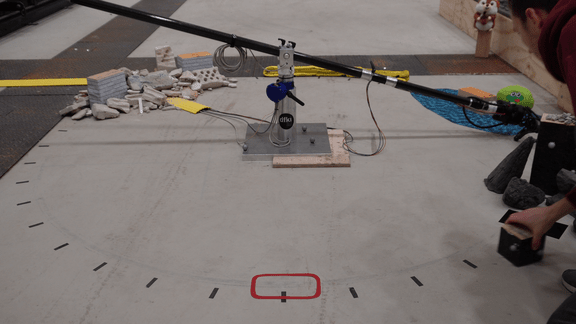}		 
\\
\includegraphics[width=\xscale]{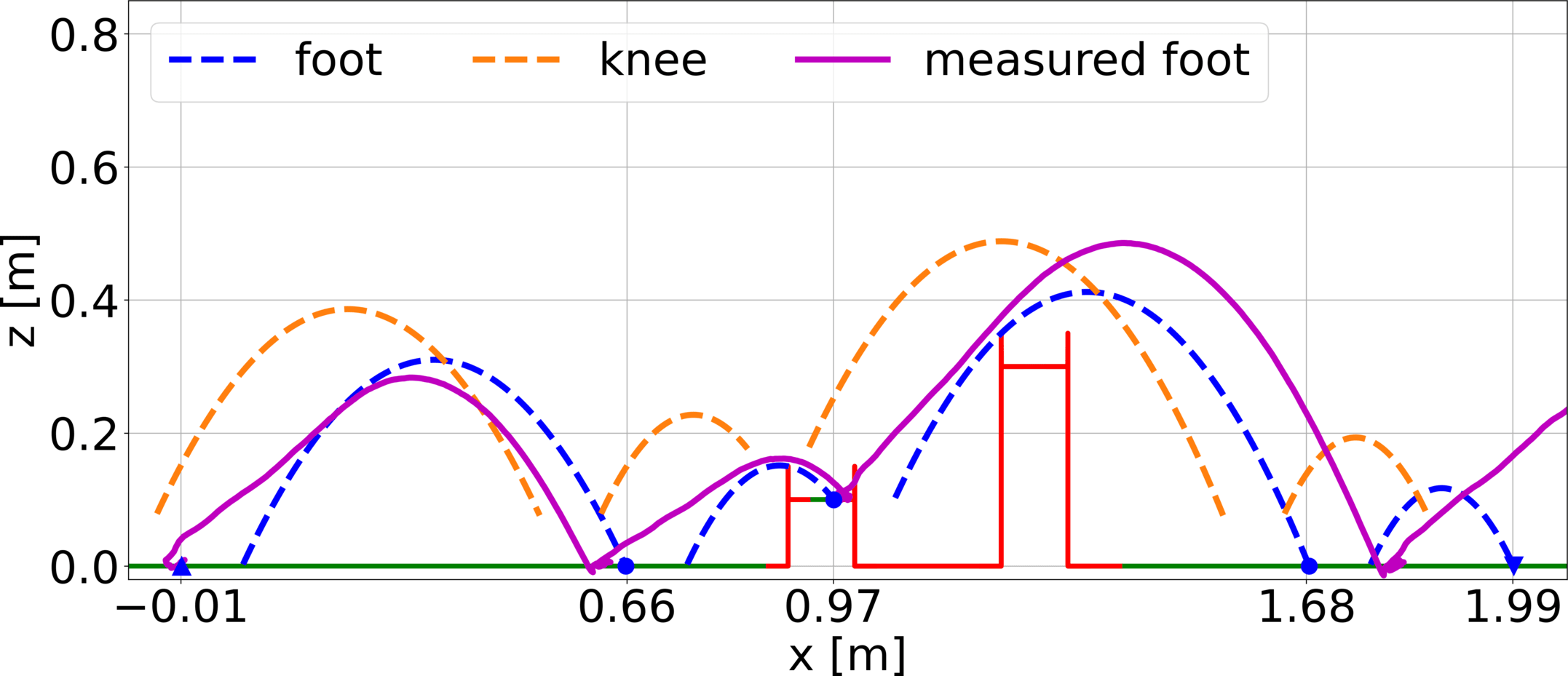}&
\includegraphics[width=\xscale]{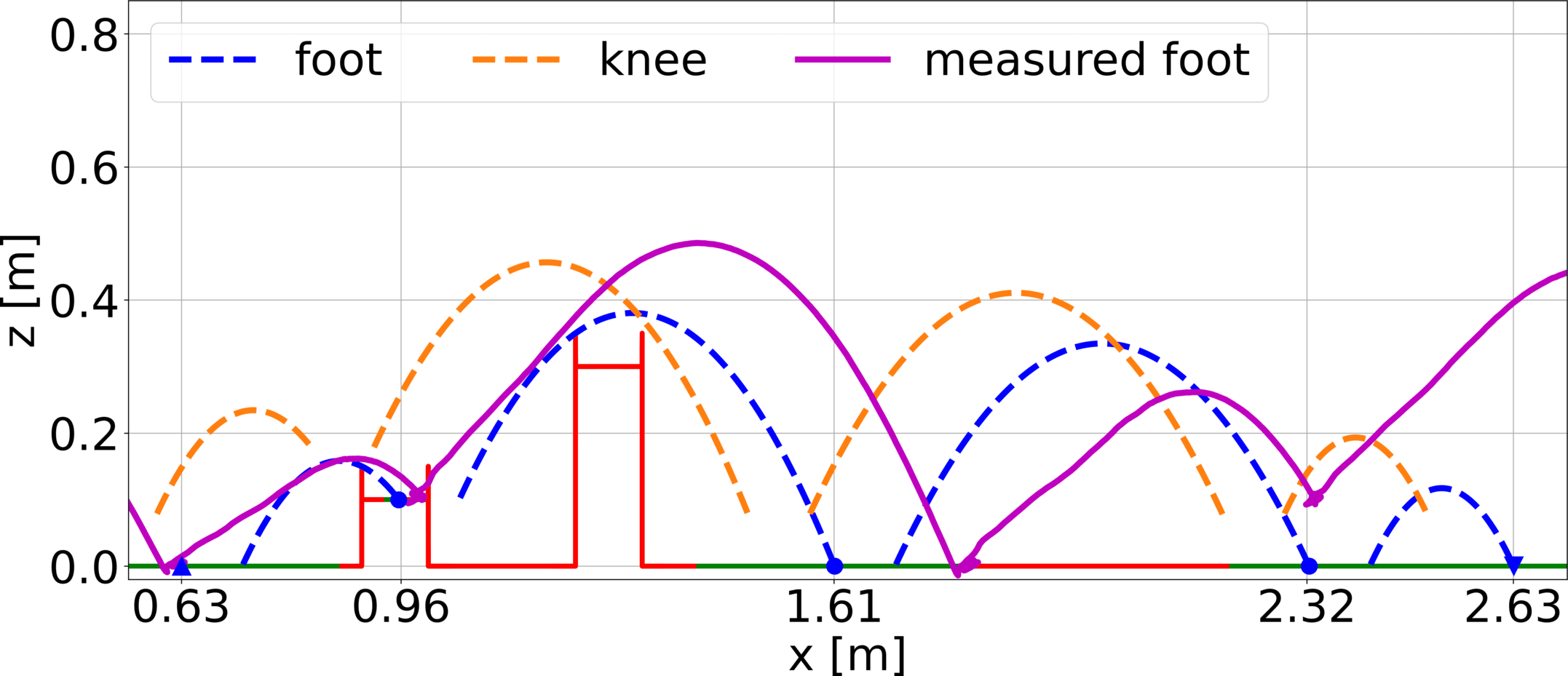}&
\includegraphics[width=\xscale]{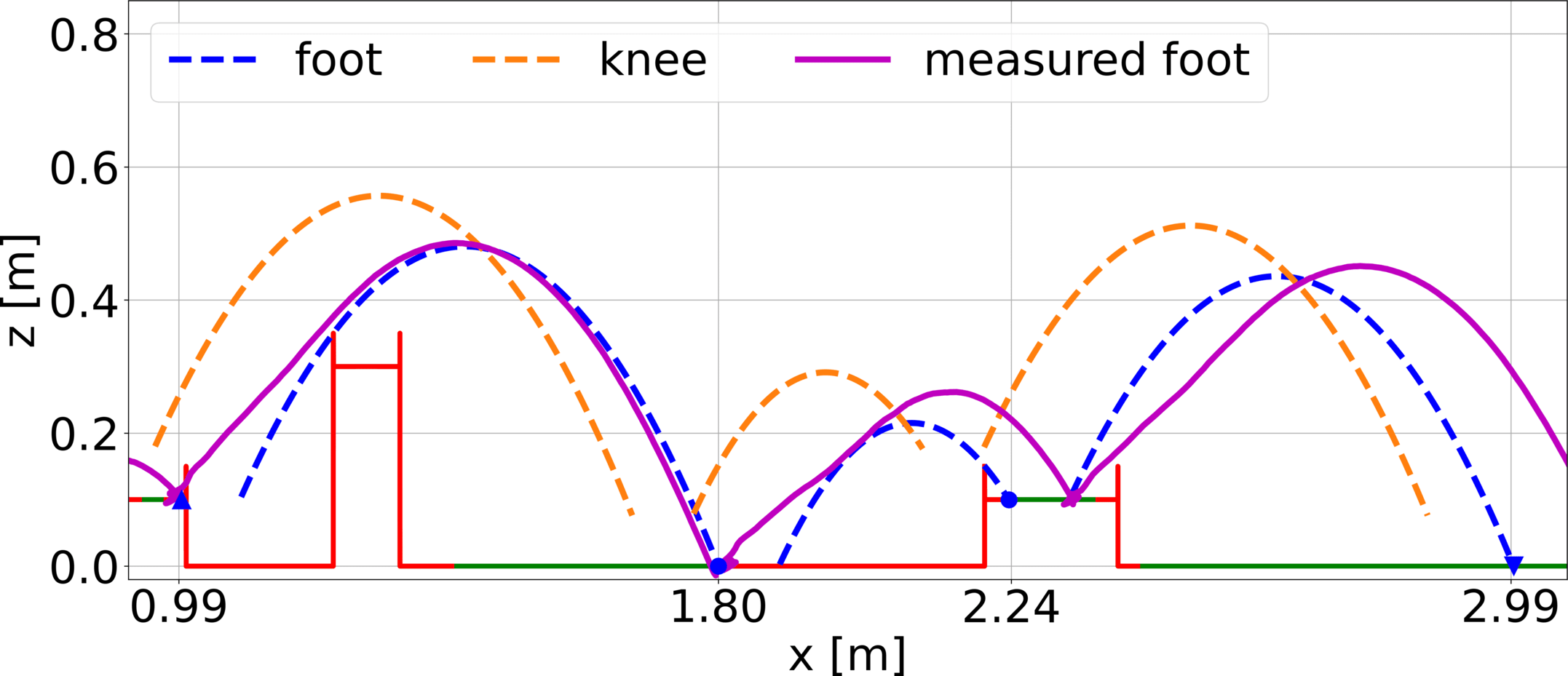}&
\includegraphics[width=\xscale]{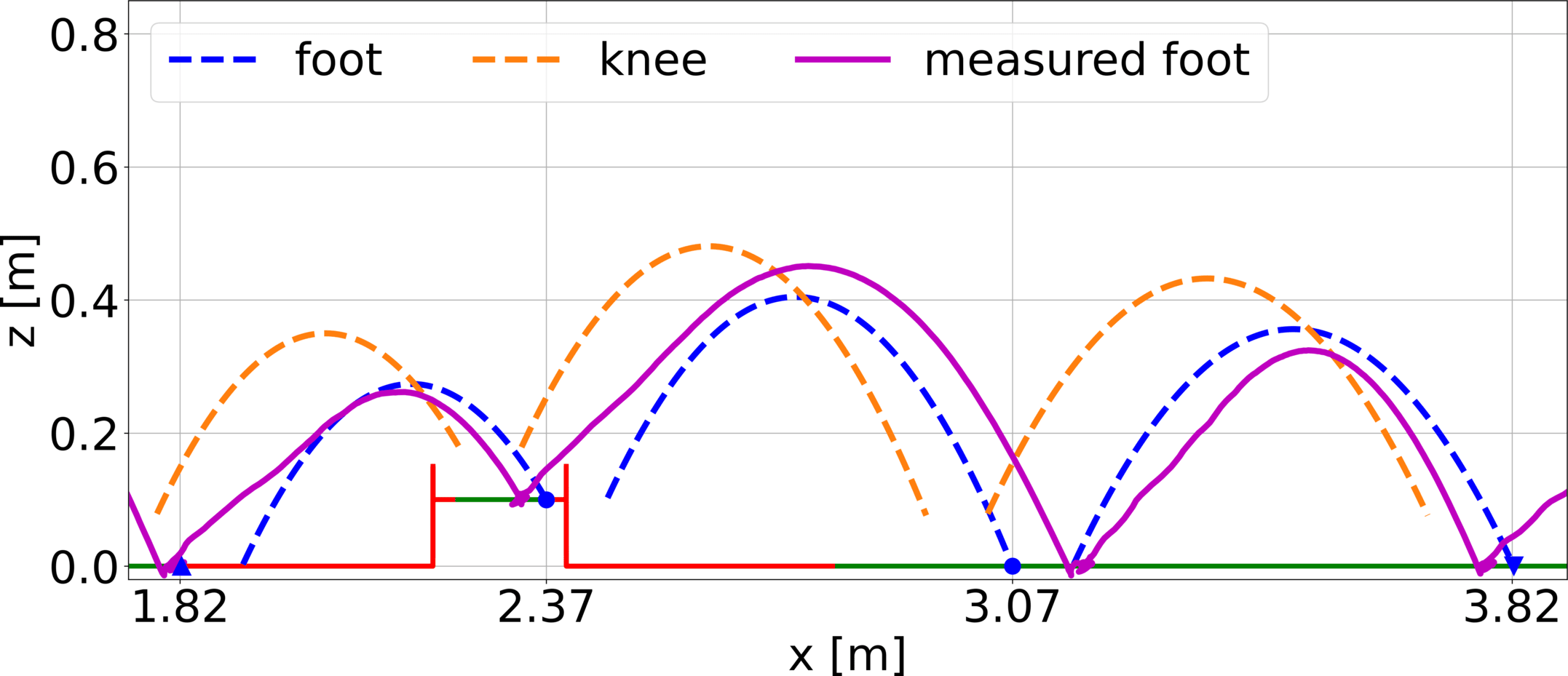}
\\
\hspace{0.01cm} & \hspace{0.01cm} & \hspace{0.01cm} & \hspace{0.01cm}
\\
\includegraphics[width=\xscale]{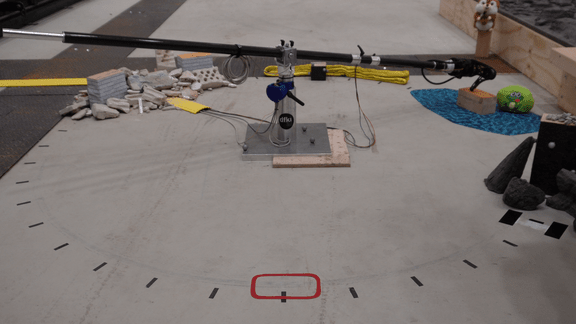}&
\includegraphics[width=\xscale]{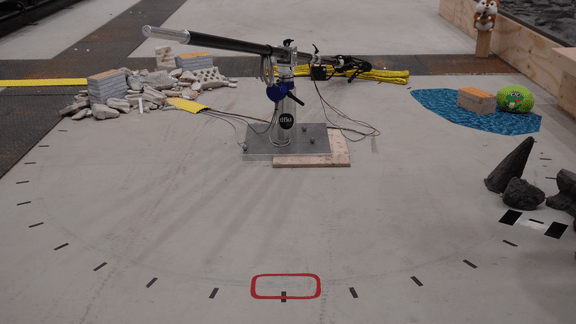}&
\includegraphics[width=\xscale]{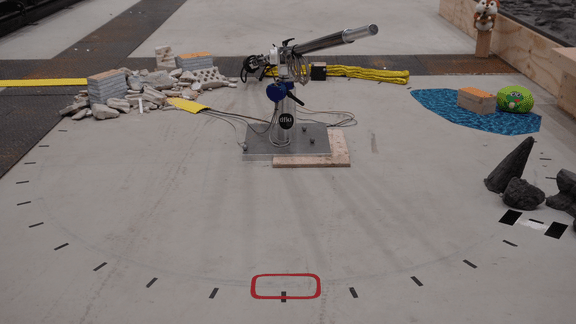}&
\includegraphics[width=\xscale]{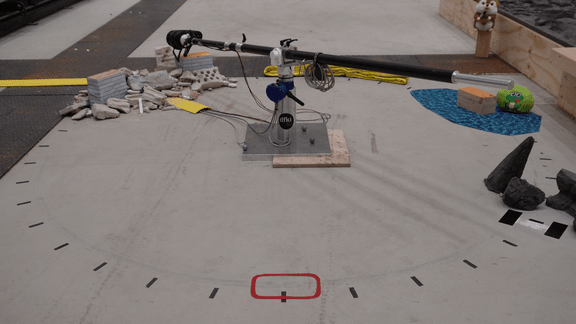}		 
\\
\includegraphics[width=\xscale]{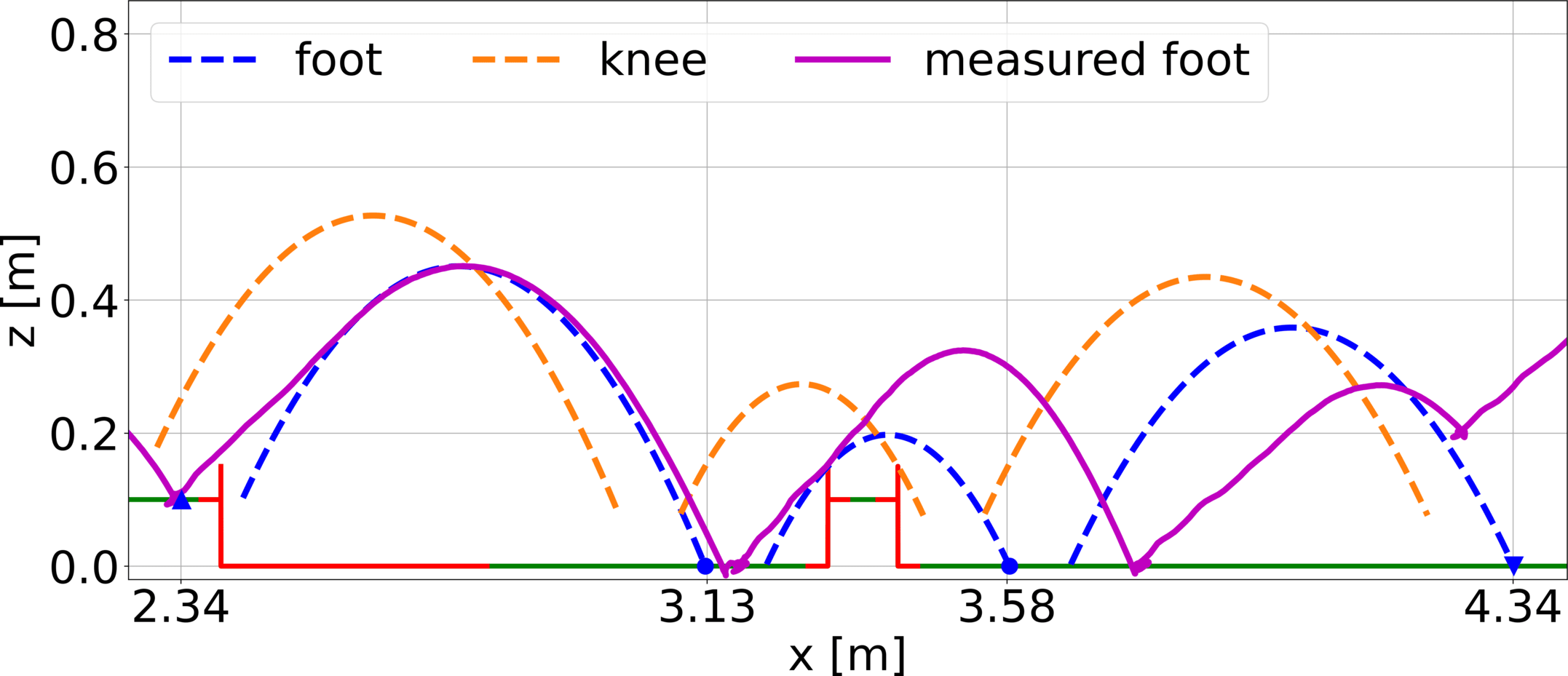}&
\includegraphics[width=\xscale]{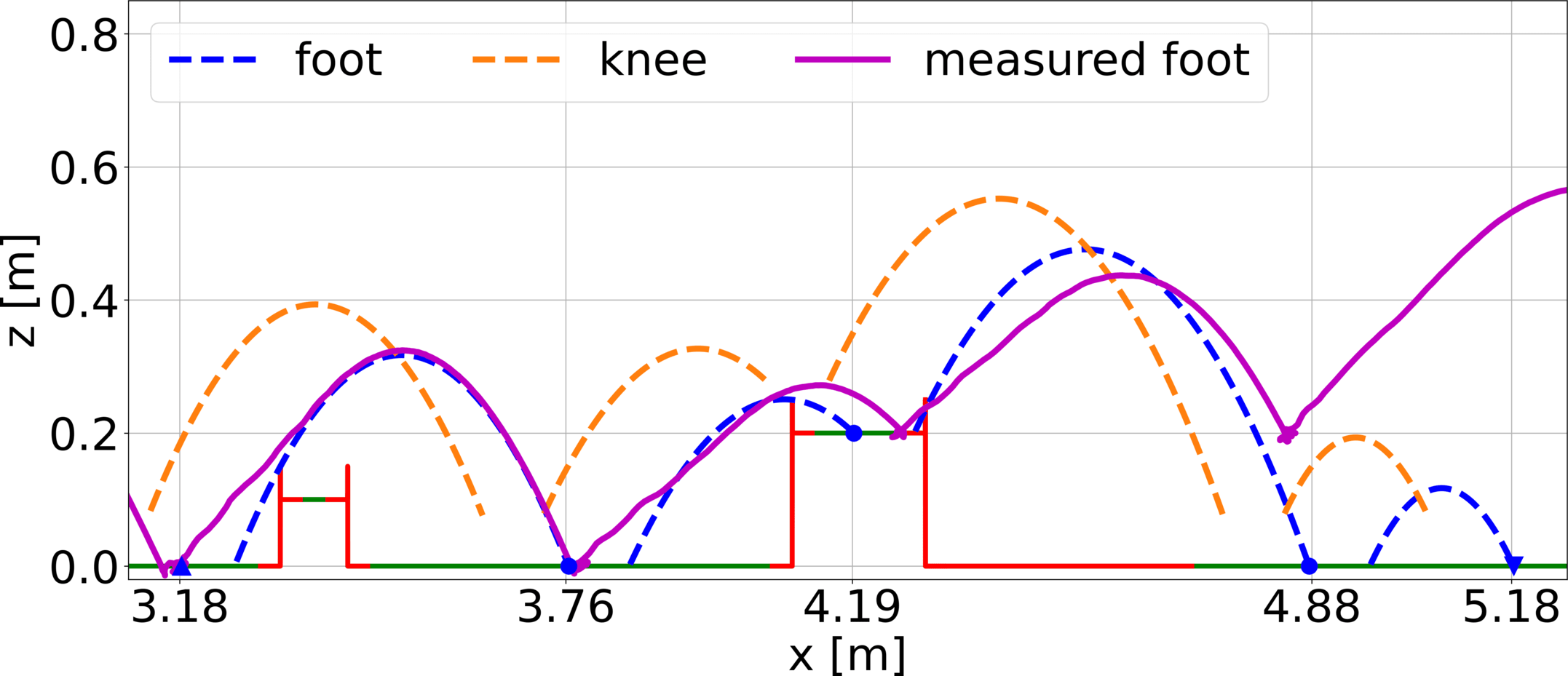}&
\includegraphics[width=\xscale]{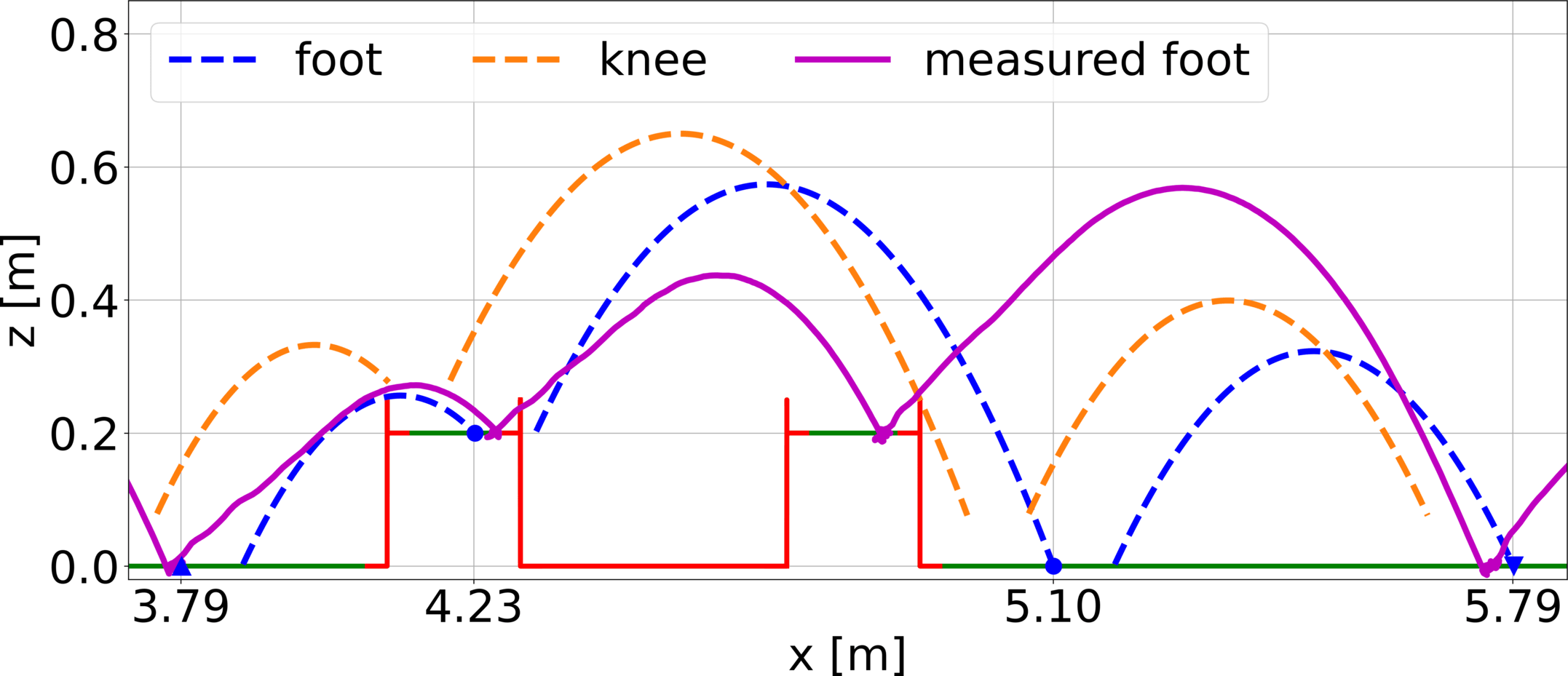}&
\includegraphics[width=\xscale]{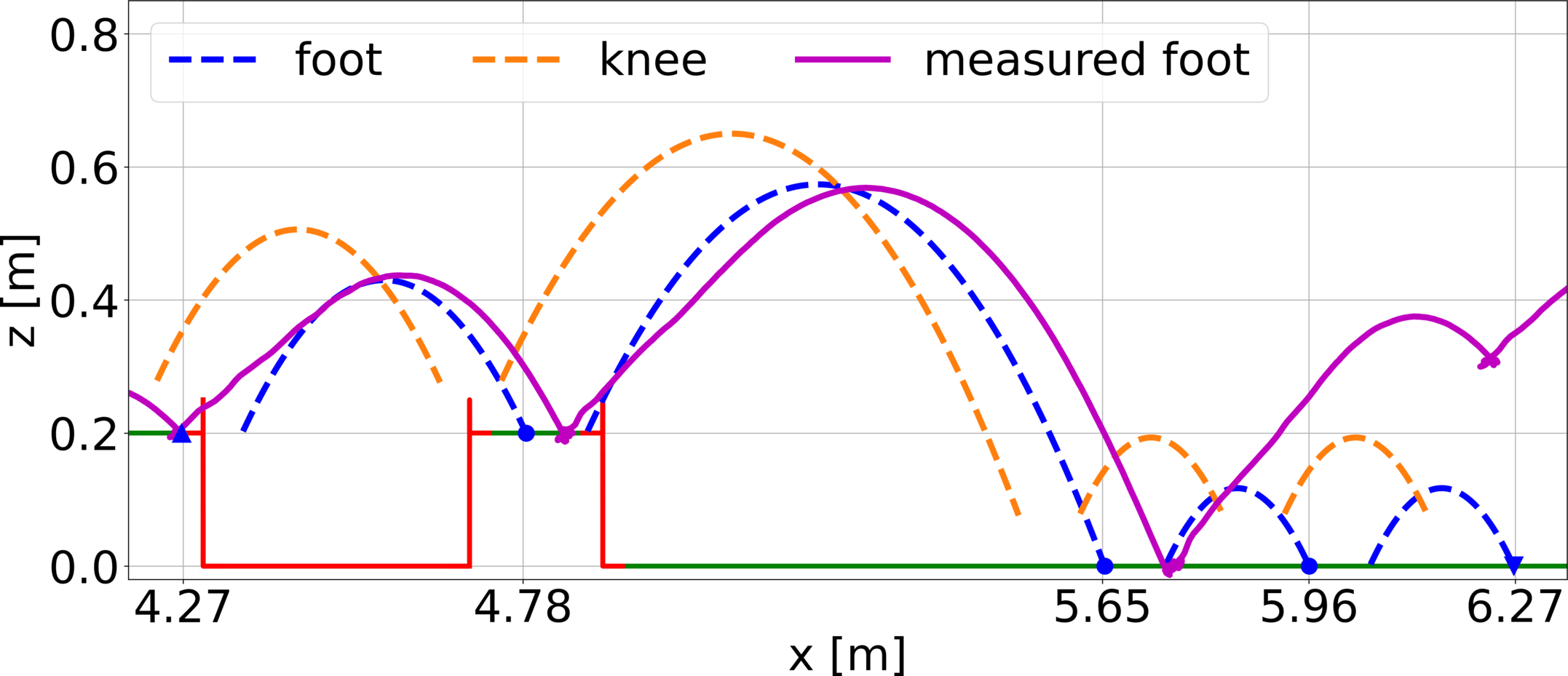}
\\
\hspace{0.01cm} & \hspace{0.01cm} & \hspace{0.01cm} & \hspace{0.01cm}
\\
\includegraphics[width=\xscale]{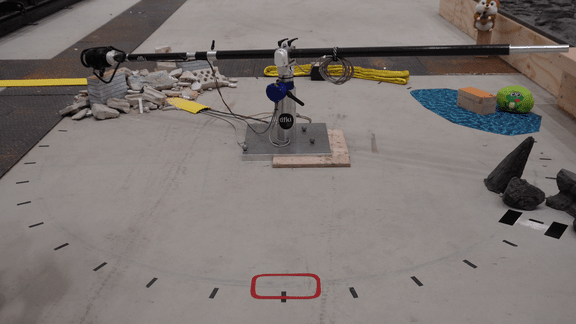}&
\includegraphics[width=\xscale]{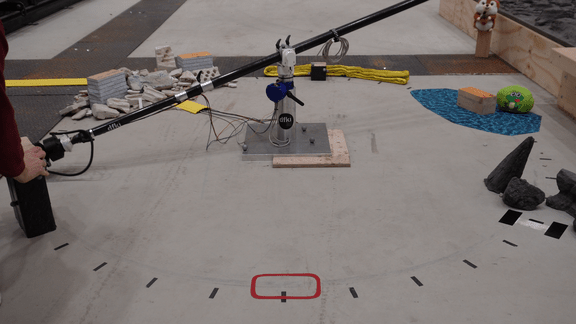}&
\includegraphics[width=\xscale]{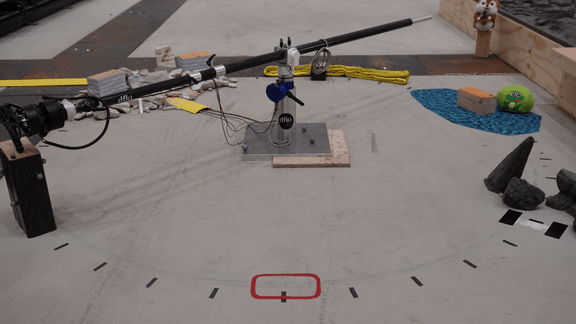}&
\includegraphics[width=\xscale]{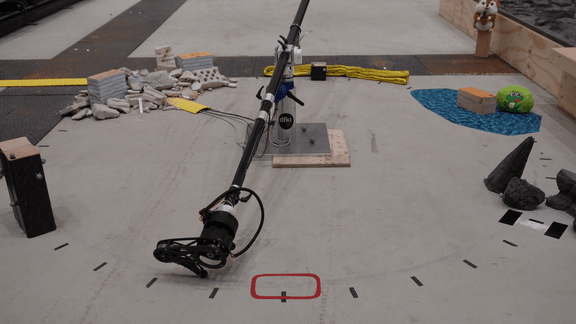}		 
\\
\includegraphics[width=\xscale]{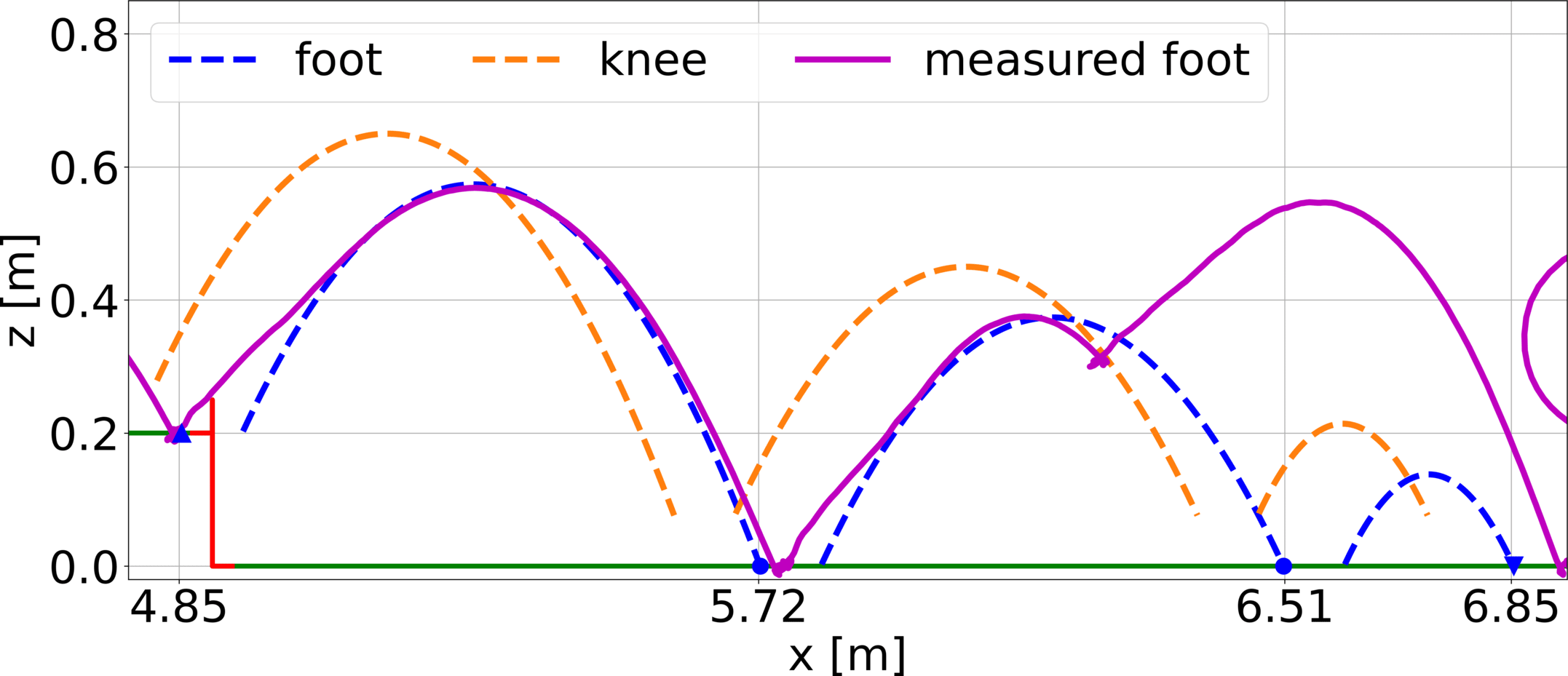}&
\includegraphics[width=\xscale]{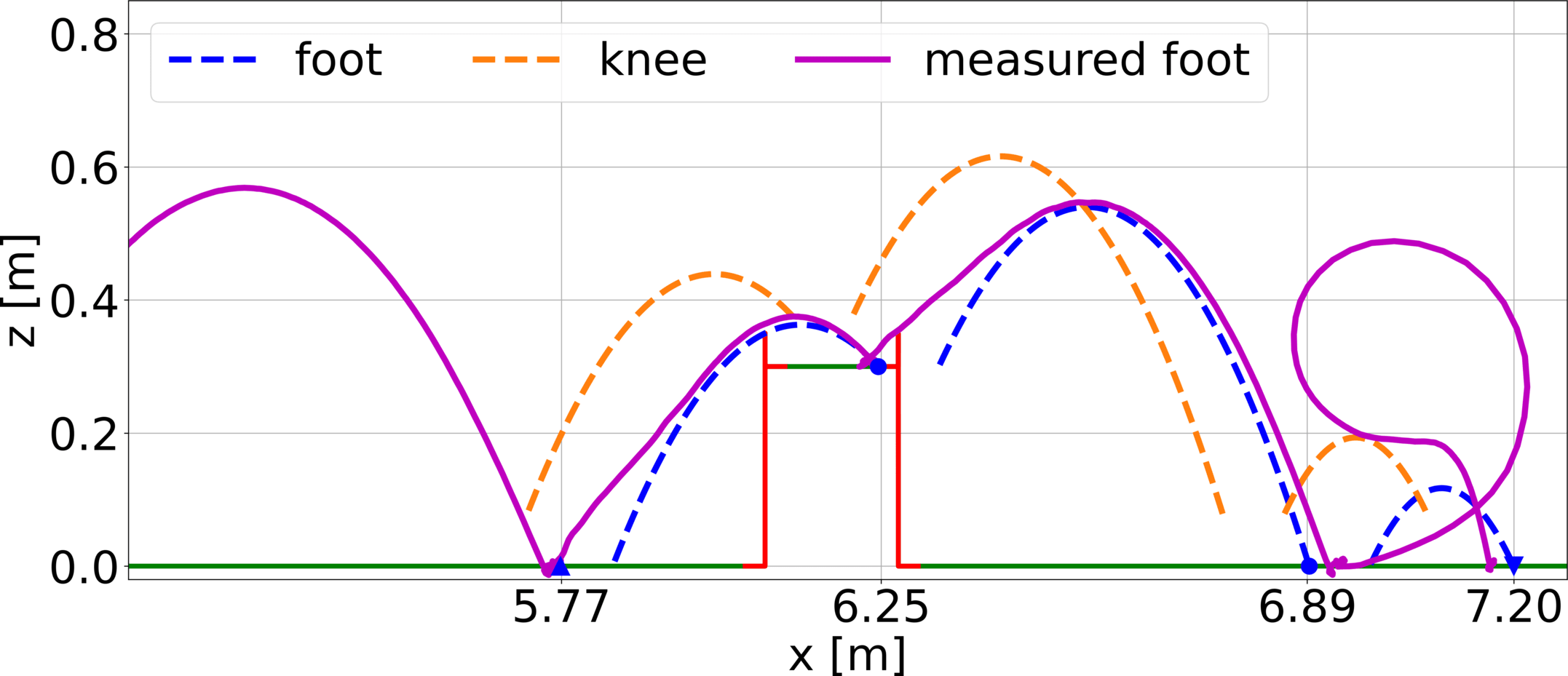}&
\includegraphics[width=\xscale]{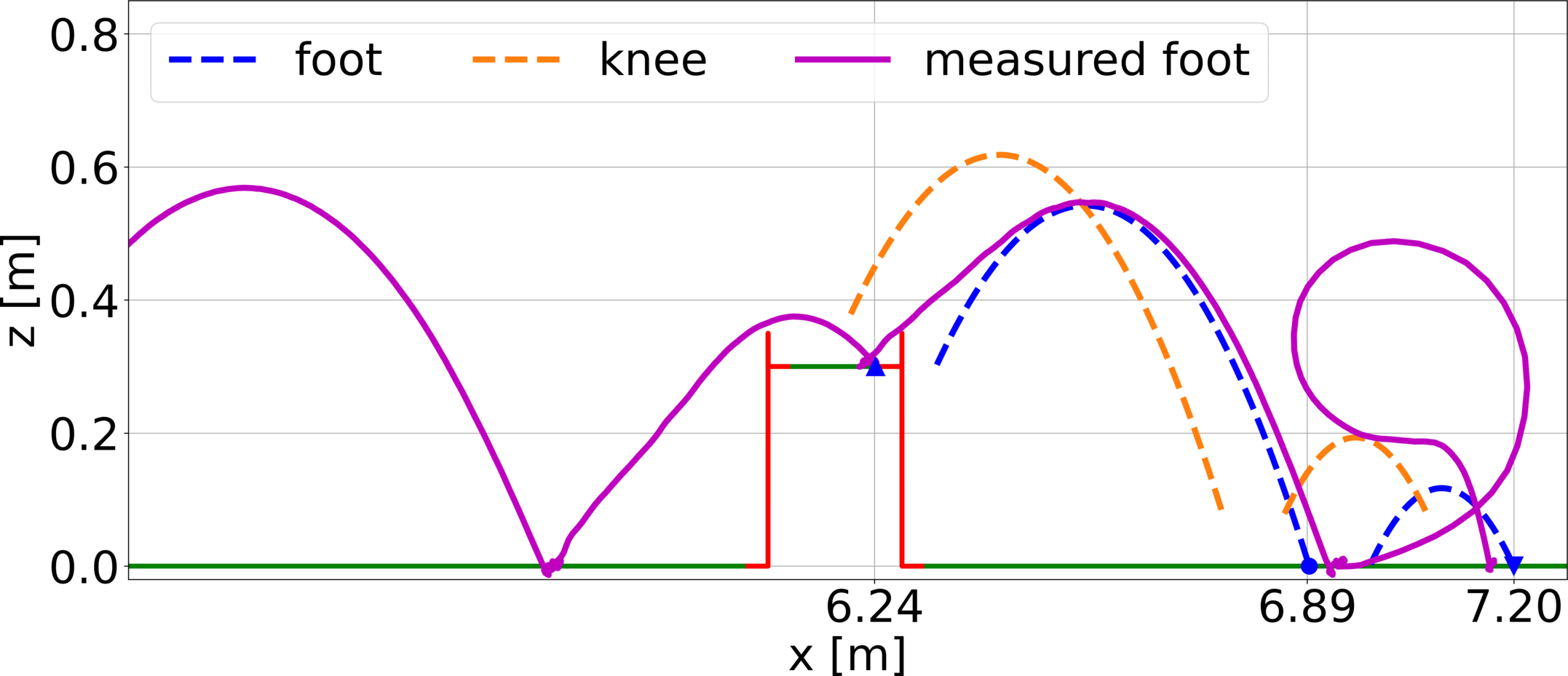}&
\includegraphics[width=\xscale]{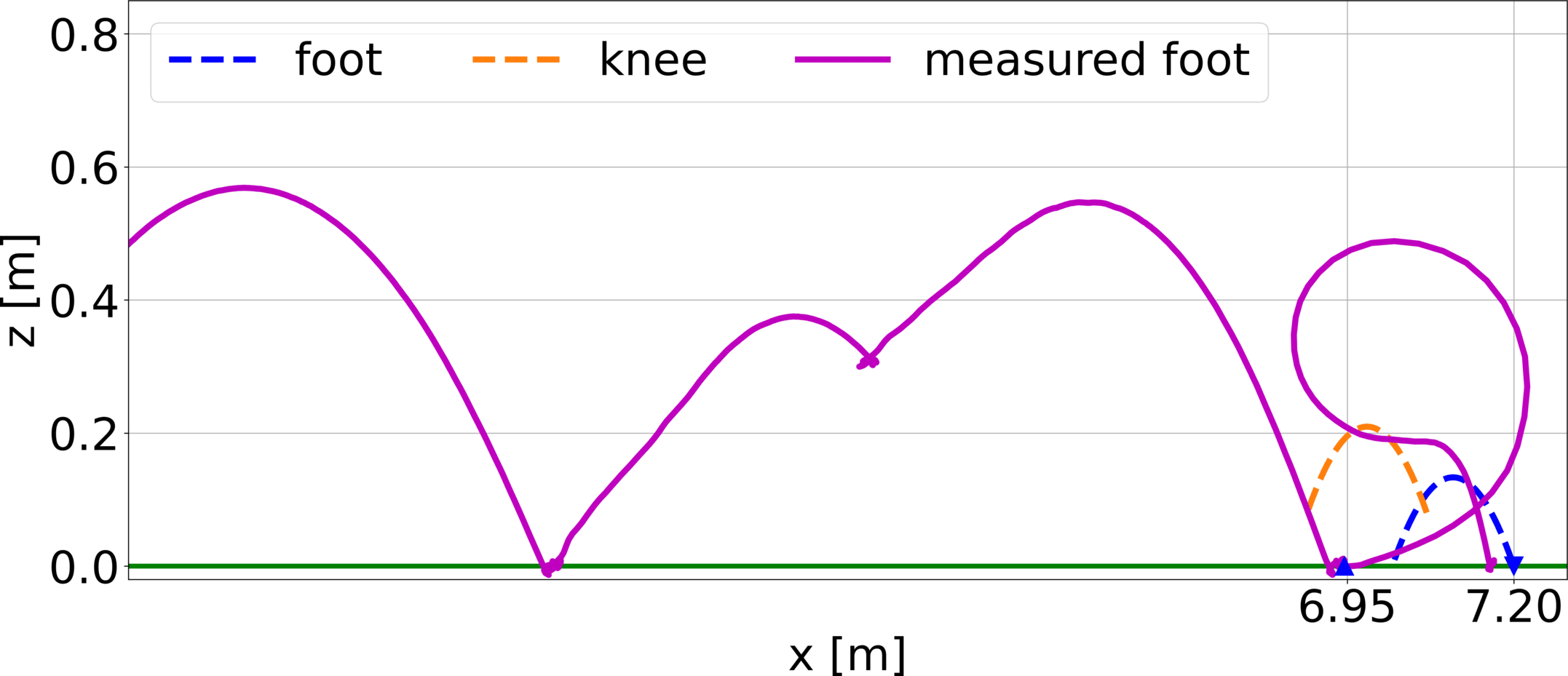}
\end{tabular}
\captionof{figure}{Snapshots of each stance and the corresponding planned and executed trajctories}
\label{fig:result_2}
\vspace{-0.4cm}
\end{table*}
\endgroup}
Only the first jump of the planned motion is carried out on the real system.
After each jump, the parkour planning is updated with new information about the system state and the obstacles. \textcolor{changedcol}{The lookahead distance $p$ is set to \SI{2}{\metre}.}
Objects that appear inside the \textcolor{changedcol}{lookahead distance} bring along new constraints that render the previous solution infeasible or suboptimal.
As a result, apart from the first jump, the actual sequence of jumps may deviate significantly from the initial plan.
It can be observed that the actual landing points do not match the planned landing points. This is attributed to model mismatches, also known as the \textit{sim-to-real gap}. However, the current position of the system is used after each jump for the re-planning of the optimal path.
This compensates for the disturbances and the robot masters the parkour in a very robust way.
The bigger tracked obstacle is placed during the execution of the ninth jump directly in front of the hopping leg.
Due to the quick and adaptive replanning, the altered obstacle course is recognized and processed, and a feasible solution is found to account for the change in the tenth jump.
The overall time for the behavior generation is always between $\SI{0.0013}{\second}$ and $\SI{0.1334}{\second}$ depending on the number of jumps and the obstacles in the \textcolor{changedcol}{lookahead distance}, as can be seen from \autoref{tab:computation}.
\begin{table}[h!]
\caption{Behaviour generation computation times}
\centering
\scalebox{1}{
\begin{tabular}{c|c|c|c}
jump & setup time [\SI{}{\milli\second}] & solve time [\SI{}{\milli\second}] & loop time [\SI{}{\milli\second}] \\
\hline
\SI{1}{} & \SI{19.6}{} & \SI{30.4}{} & \SI{72.3}{} \\
\SI{2}{} & \SI{19.9}{} & \SI{90.1}{} & \SI{134.3}{} \\
\SI{3}{} & \SI{17.0}{} & \SI{27.4}{} & \SI{45.4}{} \\
\SI{4}{} & \SI{6.5}{} & \SI{10.2}{} & \SI{17.4}{} \\
\SI{5}{} & \SI{11.3}{} & \SI{25.4}{} & \SI{37.5}{} \\
\SI{6}{} & \SI{19.8}{} & \SI{48.9}{} & \SI{93.7}{} \\
\SI{7}{} & \SI{12.4}{} & \SI{19.2}{} & \SI{32.4}{} \\
\SI{8}{} & \SI{21.1}{} & \SI{34.6}{} & \SI{79.4}{} \\
\SI{9}{} & \SI{6.1}{} & \SI{9.0}{} & \SI{15.7}{} \\
\SI{10}{} & \SI{6.1}{} & \SI{14.0}{} & \SI{27.5}{} \\
\SI{11}{} & \SI{3.1}{} & \SI{2.8}{} & \SI{6.4}{} \\
\SI{12}{} & \SI{0.5}{} & \SI{0.4}{} & \SI{1.3}{}
\end{tabular}
}
\label{tab:computation}
\vspace{-0.5cm}
\end{table}
Setup time describes the time required to set up the optimization problem and the solve time refers to the time needed by the solver to find an optimal solution.
Loop time is the overall time to find a solution including the failed tries with fewer jumps (time for the for loop in Algorithm \ref{alg:mpc}).
The behavior stabilization runs at $\SI{200}{\hertz}$.
It encounters only one timestep during each jump cycle \textcolor{changedcol}{(\autoref{statemachine})} with lower control frequency as the \textcolor{changedcol}{behavior generation} is slower (mean value of $\textcolor{changedcol}{\approx \SI{50}{\milli\second}}$).
As the behavior generation is smoothly integrated into the dynamic movement during the stance, the leg does not freeze to plan the new trajectory.
\textcolor{changedcol}{The stance phase time has a variance of $\SI{0.25}{\second}$. Therefore, behavior generation times (\autoref{tab:computation}) have an insignificant effect on the stance phase duration.}
The torque commands and measured torques for the entire experiment can be seen in \autoref{pic:torque}, summarized into the \textit{Stance} phase are the \textit{Absorption}, \textit{Reposition} and \textit{Staging} phases.
Notably in the \textit{Exertion} phase, it is mostly the knee motor contributing to the push-off force.
Peak torques appear especially in the retraction at the beginning of the \textit{Flight} phase.
\vspace{-0.4cm}
\subsection{Dynamic Environment with Disturbances}
The controller is robust to large external disturbances.
Figure \ref{fig:result_3} shows the hopping leg before and after a disturbed jump.
During flight, the end of the pole collided with an obstacle on the opposite side of the course.
As a result, the ballistic trajectory of the leg was altered and the actual landing point deviated significantly from the planned one.
As a new trajectory is calculated after each jump, such disturbances are not a reason for failure.
The run was completed despite encountering various disturbances before, including a missed landing point and a jump on slippery ground.
{\begingroup
\begin{table}[!htbp]
\setlength{\tabcolsep}{1pt}
\renewcommand{\arraystretch}{0.3}
\newcommand{\xscale}{0.49\linewidth}
\centering
\begin{tabular}{cc}
\includegraphics[width=\xscale]{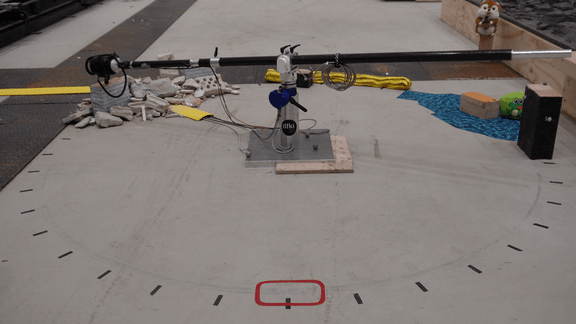}&
\includegraphics[width=\xscale]{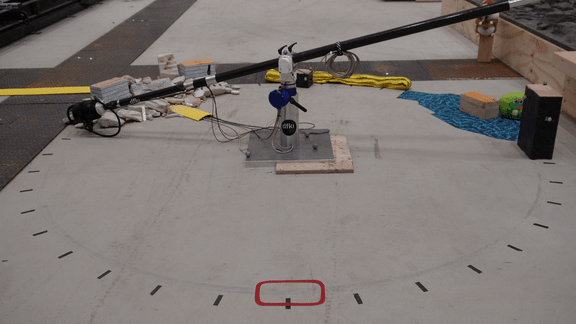}
\\
\includegraphics[width=\xscale]{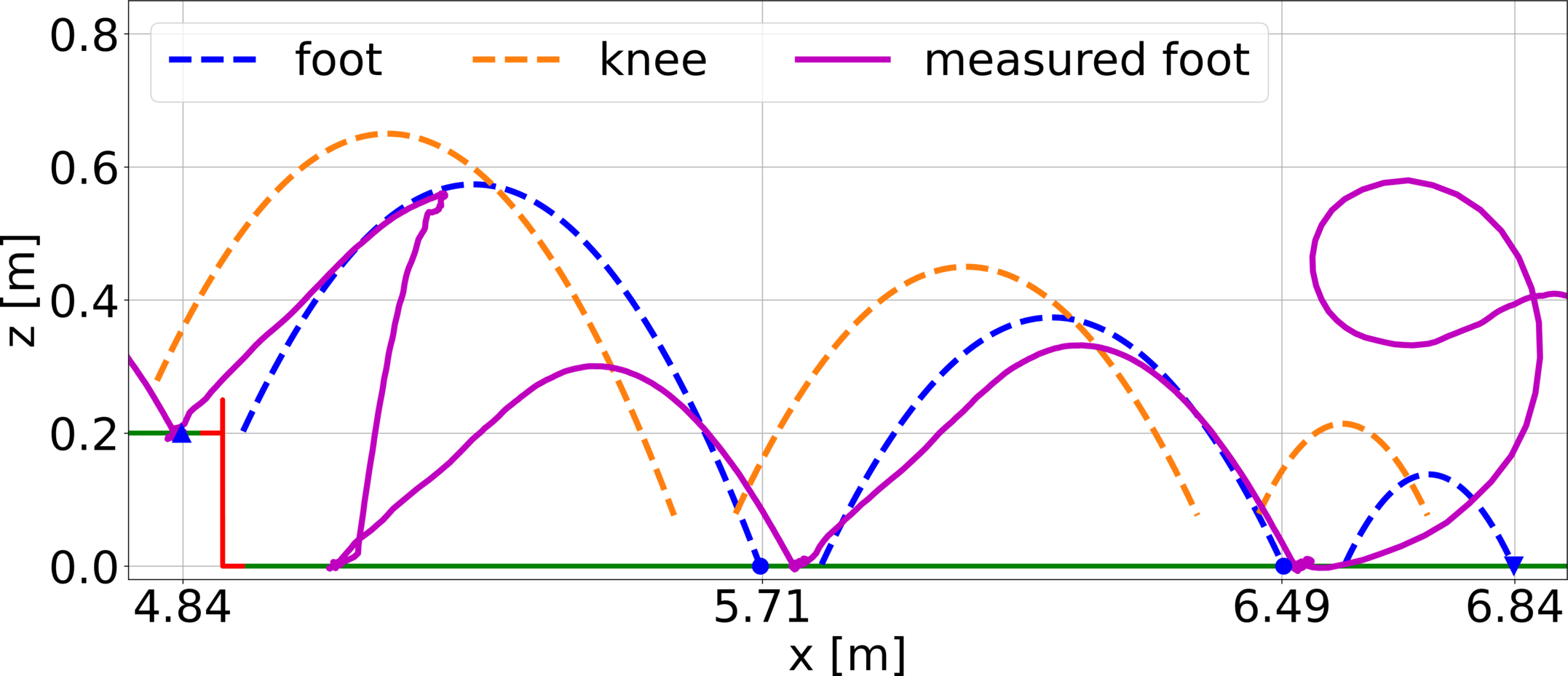}&
\includegraphics[width=\xscale]{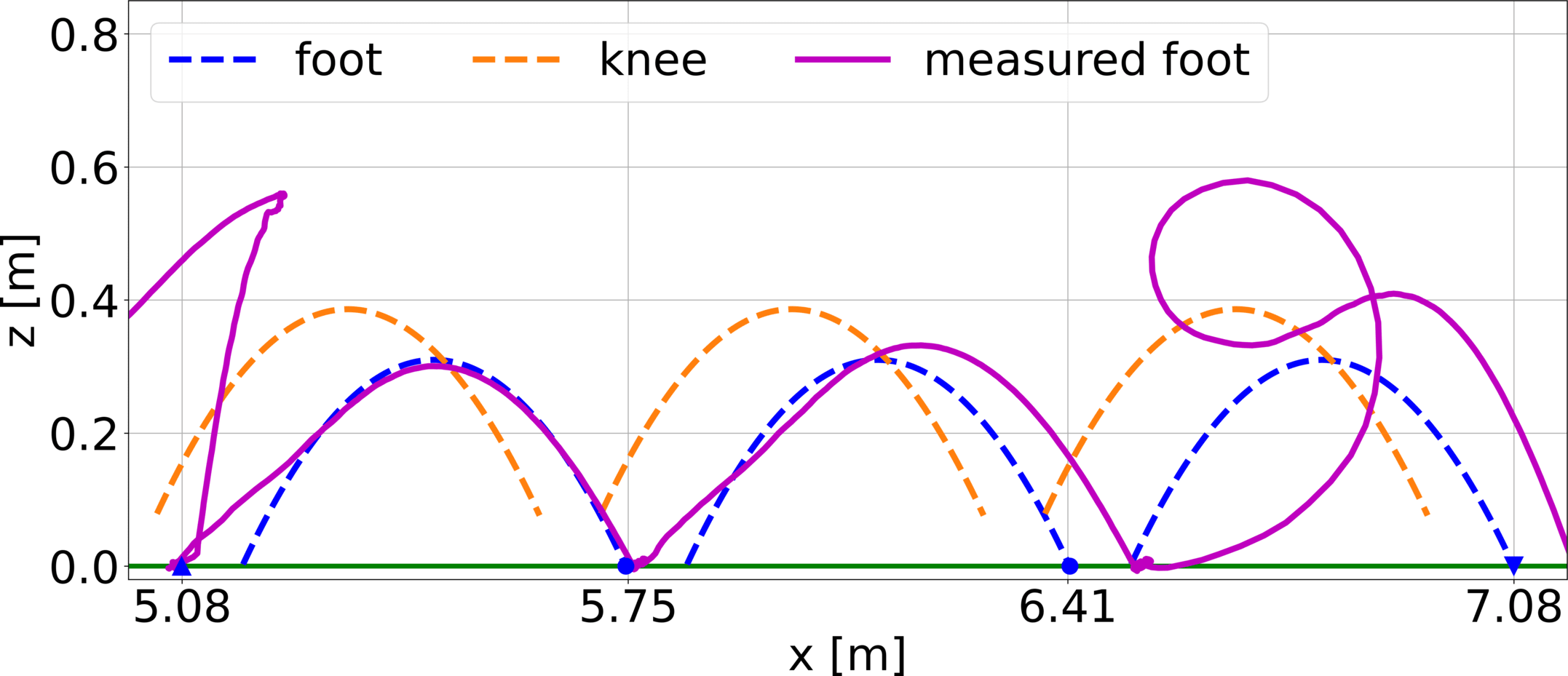}
\end{tabular}
\captionof{figure}{Snapshots before and after a disturbed jump, and corresponding planned and executed trajectories}
\label{fig:result_3}
\vspace{-0.6cm}
\end{table}
\endgroup}
\vspace{-0.3cm}
\subsection{Discussion}
\label{sec:discussion}
This receding horizon controller approach only considers a small segment of the entire environment to minimize computational load while maintaining dynamic behavior.
Consequently, jumps are likely executed which are suboptimal considering the entire course and some obstacle courses are completed with more jumps than minimally necessary.
Also contributing to this is that the solver does not successfully solve the problem for some rare instances where there would be a feasible solution available.
In those instances, a solution is still found in the following iteration, which is sufficient for the course completion but not for optimality.
However, this disadvantage does not outweigh the advantage of increased robustness.
Replanning after each jump allows large disturbances to be rejected and enables the traversal of dynamic environments with great adaptation.
The behavior stabilization framework ensures smooth and reliable jump execution with dynamic transitions between the individual states. 
\textcolor{changedcol_v2}{Similar solutions were obtained when the total effort was minimized instead of flight time as both cost functions lead to results with feasible minimum take-off angles.}


\vspace{-0.2cm}
\section{Conclusion}
\label{sec:conclusion}
Numerous successful experiments have been conducted with highly diverse parkour setups.
The implemented control system proves to be remarkably adaptive, effectively navigating through dynamically changing obstacle courses in real time.
The ability to replan after each jump is executed so swiftly that it seamlessly integrates with the robot's dynamic motion.
Despite the imprecise execution of jumps due to model discrepancies, the feedback mechanism enables robust traversal of challenging terrains over extended horizons.
Overall, the control strategy serves as a promising foundation for maneuvering legged robotic systems in highly challenging terrain, like a humanoid that moves dynamically through an obstacle course by jumping from one foot to another over and above obstacles.
Further efforts will be given to extend the MPPC to three dimensions to control free-floating legged robotic systems through parkour environments and addition of on-board perception system.


\bibliographystyle{IEEEtran}
\vspace{-0.25cm}
\bibliography{references}


 





\end{document}